\documentclass[10pt]{article} 
\usepackage[preprint]{tmlr}

\usepackage{amsmath,amsfonts,bm}

\def\eqref#1{equation~\ref{#1}}

\def\1{\bm{1}}

\DeclareMathAlphabet{\mathsfit}{\encodingdefault}{\sfdefault}{m}{sl}
\SetMathAlphabet{\mathsfit}{bold}{\encodingdefault}{\sfdefault}{bx}{n}

\usepackage{hyperref}
\usepackage{url}
\usepackage{graphicx}
\usepackage{subcaption}
\usepackage{algorithm}
\usepackage{algorithmic}
\usepackage{multirow}
\hypersetup{hidelinks}

\title{Multi-Objective Bayesian Optimization for Model Merging}

\author{\name Utkarsh Agarwal \email utkarsh.agarwal@mbzuai.ac.ae \\
\name Vamshi Bonagiri \email vamshi.bonagiri@mbzuai.ac.ae \\
\name Raul Astudillo \email raul.astudillo@mbzuai.ac.ae \\
\name Monojit Choudhury \email monojit.choudhury@mbzuai.ac.ae \\
      \addr Division of Computing and Mathematical Sciences\\
      Mohamed bin Zayed University of Artificial Intelligence
}

\def\month{MM}  
\def\year{YYYY} 
\def\openreview{\url{https://openreview.net/forum?id=XXXX}} 

\begin{document}

\maketitle

\begin{abstract}
Model merging combines trained models directly in weight space, offering a compute-efficient alternative to additional fine-tuning. Selecting merge parameters is nevertheless difficult because downstream evaluations are expensive, gradients are unavailable, and source capabilities can conflict. We formulate merge-parameter selection as a black-box multi-objective optimization problem and introduce \textbf{MOBO-Merge}, a merge-operator agnostic framework that uses multi-objective Bayesian optimization to approximate the Pareto front under a limited evaluation budget. We evaluate Qwen3-4B and Llama-3.1-8B in two-model instruction-math and three-model instruction-math-code settings using Linear, SLERP, TIES, and block-wise merge operators. On held-out benchmark partitions, MOBO-Merge obtains higher mean hypervolume than random search in 11 of 12 reported comparisons. The gain is small for one-dimensional Linear interpolation but substantially larger for several TIES, block-wise, and three-objective searches. No merge operator is uniformly best: TIES leads in three of four family-setting combinations, whereas Block-Linear 4x is strongest for the Llama three-model merge. These results show that multi-objective Bayesian optimization is valuable as a search layer for expressive merge parameterizations.
\end{abstract}

\begin{figure*}[t]
\begin{center}
\centerline{\includegraphics[width=\linewidth]{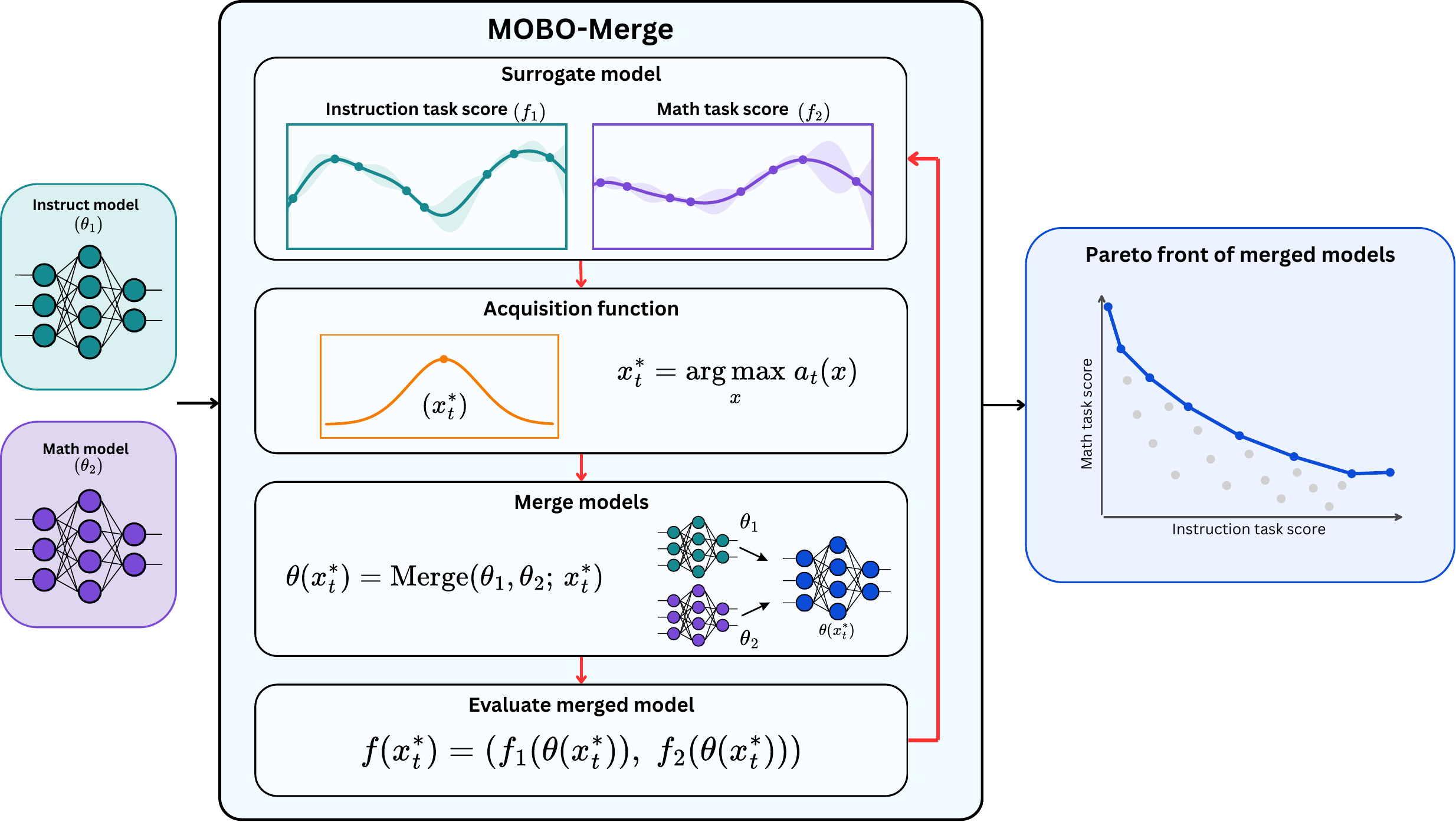}}
\caption{\textbf{Overview of MOBO-Merge.} Model merging is cast as a multi-objective optimization problem. (Left) Source models specialized for different capabilities (e.g., instruction following $\theta_1$ and mathematical reasoning $\theta_2$) to be merged via a parameterized merge operator. (Middle) Probabilistic surrogate models of the mapping from merge parameters to objective scores and are used to construct an acquisition function that selects the next merge configuration to evaluate. (Right) Iterating this process yields an approximation of the Pareto front, exposing optimal trade-offs between competing capabilities.}
\label{fig:method_overview}
\end{center}
\end{figure*}

\section{Introduction}

Model merging has emerged as a practical and increasingly popular approach for composing capabilities from pretrained and fine-tuned neural networks \citep{yang2024model}. Instead of training a single monolithic model, practitioners often fine-tune separate models for different skills—such as instruction following, mathematical reasoning, or code generation—and then combine these specialized models into a single system. This paradigm enables modular reuse of existing models while avoiding additional training, and has led to strong empirical results across a wide range of settings \citep{wortsman2022model, ilharco2022editing, yadav2023ties, yu2024language, goddard2024arcee}.

Most model merging methods operate directly in weight space using relatively simple operations. For example, given two models $\theta_1$ and $\theta_2$, a common strategy is linear interpolation,
\[
\theta(x) = x \theta_1 + (1 - x) \theta_2,
\]
where $x \in [0,1]$ controls the relative contribution of each model \citep{wortsman2022model, ilharco2022editing}. Even in this simple case, downstream performance can vary sharply as a function of the merge weight, and small changes can lead to qualitatively different behavior. In practice, many merging methods employ richer parameterizations—such as block-wise mixing \citep{yuce2025kafa} or interference-aware operators \citep{yadav2023ties}—which induce low- to moderate-dimensional continuous merge spaces in which naive search strategies quickly become impractical \citep{yuce2025kafa,du2024parameter}.

Crucially, model merging is often inherently multi-objective. Source models are typically optimized for different capabilities and evaluated using distinct benchmarks, so improving performance on one capability frequently degrades performance on another. As a result, no single merge configuration performs optimally across all criteria \citep{li2025s, chen2025bambo, li2024map}. Model merging is therefore best understood in terms of trade-offs rather than a single optimal solution, with solution quality naturally characterized by the Pareto front it induces.

Identifying high-quality merges under these trade-offs is challenging. Each candidate merge must be instantiated and evaluated on costly downstream benchmarks; evaluations are often noisy, gradients with respect to merge parameters are unavailable, and evaluation budgets are limited. Consequently, exhaustive grid search is infeasible, while random or heuristic strategies scale poorly as the merge space grows. These characteristics place model merging squarely in the regime of expensive black-box multi-objective optimization.

In this paper, we introduce \textbf{Multi-Objective Bayesian Optimization for Model Merging (MOBO-Merge)}, a framework that casts merge-weight selection as a black-box multi-objective optimization problem and applies multi-objective Bayesian optimization \citep{knowles2006parego,daulton2020differentiable} to efficiently approximate the Pareto front of capability trade-offs. Rather than targeting a single merged model, MOBO-Merge explicitly seeks a diverse set of Pareto-optimal merges, enabling practitioners to select models that best match their desired balance of capabilities under fixed compute budgets. Importantly, the framework is agnostic to the choice of merge operator and applies uniformly across global interpolation, interference-aware merging, and higher-dimensional block-wise parameterizations.

We evaluate MOBO-Merge on Qwen3-4B \citep{qwen3_4b} and Llama-3.1-8B \citep{llama3_8b}, using two-model instruction-math merges and three-model instruction-math-code merges. Across held-out comparisons, MOBO-Merge obtains higher mean hypervolume than random search in 11 of 12 settings. The gain is more pronounced for several TIES, block-wise, and three-objective searches. The experiments also show that no merge operator is uniformly best: TIES leads in three of four family-setting combinations, while Block-Linear 4x performs best for the Llama three-model merge.
Across experiments, MOBO-Merge reliably recovers high-quality Pareto fronts with substantially fewer evaluations than grid or random search and exposes controllable capability trade-offs that are difficult to uncover with single-objective or heuristic approaches. Beyond these gains, the generality of the framework enables a systematic study of multi-objective optimization across diverse merging strategies. This analysis yields several practical insights, including that interference-aware methods such as TIES are relatively robust to their merge weights, and that higher-dimensional block-wise linear merges can achieve competitive performance while remaining simple and interpretable. Together, these findings suggest that efficiently exploring richer merge parameterizations can unlock strong performance trade-offs even for simple merging rules.

\paragraph{Contributions.}
\begin{itemize}
   \item We formulate merge-parameter selection for specialized models as an
expensive black-box multi-objective optimization problem and present \textbf{MOBO-Merge}, a merge-operator-agnostic framework that uses multi-objective Bayesian optimization to efficiently approximate sets of
Pareto-optimal merges under limited evaluation budgets.
    \item We evaluate two contemporary model families in two-model and three-model settings across global, interference-aware, and block-wise merge operators, with deterministic validation/held-out benchmark splits.
    \item We show that the advantage over random search grows in several expressive merge spaces, while also identifying important trends: Linear search is already competitive in one dimension, and the strongest merge operator depends on the model family and objective set.
\end{itemize}

\section{Problem Formulation}

We formulate model merging as a parameterized decision problem in which a small number of continuous merge parameters control how multiple trained models are combined, and performance is assessed through costly downstream evaluations. Rather than treating merging as a fixed procedure, we view it as a mapping from a low-dimensional merge space to a vector of task-level outcomes.

Let $\Theta$ denote the parameter space of a shared model architecture, and let $\theta_1, \dots, \theta_K \in \Theta$ denote trained models to be merged. These models may be pretrained or fine-tuned for different purposes but are assumed to share a common parameterization.

Merging is specified through a \emph{merge operator}. Let $\mathcal{X} \subset \mathbb{R}^D$ denote a low-dimensional merge parameter space encoding how the source models are combined. The merge operator
\[
\mathcal{M} : \Theta^K \times \mathcal{X} \rightarrow \Theta
\]
produces a merged model
\[
\theta(\boldsymbol{x}) = \mathcal{M}(\theta_1, \dots, \theta_K; \boldsymbol{x}),
\]
parameterized by $\boldsymbol{x} \in \mathcal{X}$.

This abstraction treats the merge operator as a black box and places minimal assumptions on its structure. The merge space $\mathcal{X}$ may represent global mixture weights, block-wise coefficients, or other structured controls, allowing the formulation to capture a wide range of existing and future merging strategies.

Merged models are evaluated using a collection of objective functions
\[
f_i : \Theta \rightarrow \mathbb{R}, \quad i = 1, \dots, M,
\]
where each $f_i$ measures performance with respect to a particular capability or evaluation criterion. These objectives are typically defined by external benchmarks and may be expensive or noisy to evaluate.

The merge operator induces a vector-valued objective function over the merge space,
\[
\boldsymbol{f}(\boldsymbol{x}) = \big( f_1(\theta(\boldsymbol{x})), \dots, f_M(\theta(\boldsymbol{x})) \big),
\]
mapping merge parameters directly to task-level outcomes.

In many practical settings, the objectives correspond to distinct capabilities and are inherently conflicting: improving performance on one objective often degrades performance on another. Consequently, we do not assume the existence of a single optimal merge. Instead, solution quality is defined in terms of \emph{Pareto optimality} \citep{daulton2020differentiable}.

A merge parameter $\boldsymbol{x} \in \mathcal{X}$ is Pareto-optimal if there is no $\boldsymbol{x}' \in \mathcal{X}$ such that
\[
\boldsymbol{f}(\boldsymbol{x}') \succeq \boldsymbol{f}(\boldsymbol{x})
\quad \text{and} \quad
\boldsymbol{f}(\boldsymbol{x}') \neq \boldsymbol{f}(\boldsymbol{x}),
\]
where $\succeq$ denotes component-wise comparison. The set of Pareto-optimal parameters defines the \emph{Pareto set}, and its image under $\boldsymbol{f}$ forms the \emph{Pareto front}. In practice, the goal is to efficiently approximate this Pareto front under a limited evaluation budget.

Evaluating $\boldsymbol{f}(\boldsymbol{x})$ requires instantiating a merged model and running downstream benchmarks. Evaluations are costly, may be noisy, and provide no gradients with respect to $\boldsymbol{x}$, so only a small number of evaluations is feasible. These properties place model merging squarely in the regime of expensive black-box multi-objective optimization.

\subsection{The Canonical Case: Merging Specialized Models}

A common instantiation of the formulation above occurs when each source model is trained or fine-tuned to specialize in a distinct capability. In this case, the objectives often correspond directly to these capabilities, and the number of objectives matches the number of source models, i.e., $M = K$ \citep{li2025s, chen2025bambo}.

Each source model $\theta_k$ is therefore approximately optimized for its corresponding objective $f_k$, but is typically suboptimal with respect to the remaining objectives. Because the objectives are inherently conflicting, improving performance on one capability often comes at the expense of others. Consequently, no single merge is expected to dominate across all objectives, and the set of desirable solutions is naturally characterized by a Pareto front.

While our experiments focus on this canonical setting, the formulation itself is more general. In particular, the number of objectives need not match the number of source models, and objectives may represent arbitrary evaluation criteria beyond model-specific capabilities.

\section{Method: MOBO-Merge}

For completeness, we describe the standard multi-objective Bayesian
optimization procedure used by \textbf{MOBO-Merge}. Given a parameterized
merge operator, MOBO-Merge treats the induced vector-valued objective
function $f:\mathcal{X}\rightarrow\mathbb{R}^{M}$ as an expensive black box
and uses probabilistic surrogate models and a hypervolume-based acquisition
function to approximate its Pareto front under a limited evaluation budget.
The framework is merge-operator agnostic: the same optimization procedure
applies whenever a merge operator defines a parameter space $\mathcal{X}$ and
the resulting merged models can be evaluated on the objectives. We present
the fully sequential setting, in which one merge is selected and evaluated at
each iteration; batch extensions follow standard multi-objective Bayesian
optimization practice.

\paragraph{Overview.}
At iteration $t$, MOBO-Merge maintains a probabilistic surrogate model of the objectives based on previously evaluated merge parameters. This surrogate is used to construct an acquisition function that scores candidate merge configurations according to their expected contribution to Pareto-front quality. The highest-scoring configuration is selected, evaluated to obtain new objective values, and incorporated into the dataset used to update the surrogate model. Repeating this process under a fixed evaluation budget yields a set of evaluated merge configurations that together form an approximation of the Pareto front. MOBO-Merge is summarized in Algorithm~\ref{alg:mobomerge}.

\subsection{Probabilistic Modeling of the Objectives}

MOBO-Merge maintains a probabilistic model over the unknown objective function $\boldsymbol{f}$. Given a dataset
\[
\mathcal{D}_t = \left\{ \left(\boldsymbol{x}_j, \boldsymbol{y}_j \right) \right\}_{j=1}^t,
\]
where $\boldsymbol{y}_j$ denotes a (potentially noisy) evaluation of the latent objective values $\boldsymbol{f}(\boldsymbol{x}_j)$, the surrogate model induces a posterior distribution over $\boldsymbol{f}$, and in particular over $\boldsymbol{f}(\boldsymbol{x})$ for any $\boldsymbol{x} \in \mathcal{X}$. This posterior captures both epistemic uncertainty about the objective functions and observation noise in the evaluations, and is used to guide the selection of future merge parameters.

In general, any probabilistic regression model suitable for expensive black-box optimization may be used, provided it yields uncertainty-aware predictions. In our experiments, we instantiate this model using independent Gaussian process (GP) surrogates for each objective, a standard choice in Bayesian optimization \citep{frazier2018tutorial,garnett2023bayesian}

\subsection{Acquisition Function}

Given the posterior predictive distribution
$p\!\left( \boldsymbol{f}(\boldsymbol{x}) \mid \mathcal{D}_t \right)$,
MOBO-Merge constructs a multi-objective acquisition function
\[
a_t : \mathcal{X} \rightarrow \mathbb{R},
\]
which scores candidate merge parameters by their expected improvement to the current Pareto-front approximation. In the fully sequential setting considered here, the acquisition function is optimized to select a single merge parameter at each iteration.

To directly target Pareto-front quality under a limited evaluation budget, we focus on acquisition functions based on dominated hypervolume, a standard scalar measure of Pareto-front quality. While MOBO-Merge is compatible with a variety of multi-objective acquisition strategies, we use (noisy) Expected Hypervolume Improvement (NEHVI) in our experiments \citep{daulton2021parallel}.

\paragraph{Noisy Expected Hypervolume Improvement.}
Let $\boldsymbol{r} \in \mathbb{R}^M$ denote a reference point dominated by all objective values of interest. The dominated hypervolume of a set $\mathcal{S} \subset \mathbb{R}^M$ is defined as
\[
\mathrm{HV}(\mathcal{S}) = \lambda\!\left(
\bigcup_{\boldsymbol{y} \in \mathcal{S}} [r_1, y_1] \times \cdots \times [r_M, y_M]
\right),
\]
where $\lambda(\cdot)$ denotes the Lebesgue measure.

At iteration $t$, let $\mathcal{X}_t = \{\boldsymbol{x}_1,\dots,\boldsymbol{x}_t\}$ denote the set of evaluated merge parameters, and let $\boldsymbol{f}(\mathcal{X}_t)$ denote the corresponding (latent) objective values. For a candidate merge parameter $\boldsymbol{x} \in \mathcal{X}$, the noisy expected hypervolume improvement is defined as
\[
a_t(\boldsymbol{x})
= \mathbb{E}\!\left[
\mathrm{HV}\!\left(\boldsymbol{f}(\mathcal{X}_t) \cup \{\boldsymbol{f}(\boldsymbol{x})\}\right)
- \mathrm{HV}\!\left(\boldsymbol{f}(\mathcal{X}_t)\right)
\;\middle|\; \mathcal{D}_t
\right],
\]
where the expectation is taken with respect to the joint posterior distribution over $\boldsymbol{f}(\mathcal{X}_t)$ and $\boldsymbol{f}(\boldsymbol{x})$ conditioned on the observed data $\mathcal{D}_t$. Although we focus on the fully sequential setting, NEHVI naturally extends to batch selection \citep{daulton2021parallel}.

\subsection{Acquisition Optimization and Evaluation}

At iteration $t+1$, MOBO-Merge selects the next merge parameters by solving
\[
\boldsymbol{x}_{t+1} \in \arg\max_{\boldsymbol{x} \in \mathcal{X}} a_t(\boldsymbol{x}),
\]
using a suitable continuous optimization method. The selected parameters $\boldsymbol{x}_{t+1}$ are then evaluated by instantiating the merged model $\theta(\boldsymbol{x}_{t+1})$ and computing the corresponding objective values
$\boldsymbol{y}_{t+1} = \boldsymbol{f}(\boldsymbol{x}_{t+1})$.
The dataset is updated as
\[
\mathcal{D}_{t+1}
= \mathcal{D}_t \cup \{(\boldsymbol{x}_{t+1}, \boldsymbol{y}_{t+1})\},
\]
and the probabilistic surrogate model is refit.

\begin{algorithm}[t]
\caption{MOBO-Merge}
\label{alg:mobomerge}

\begingroup
\let\AND\relax
\begin{algorithmic}[1]
\REQUIRE Merge parameter space $\mathcal{X}$, evaluation budget $T$
\STATE Initialize dataset $\mathcal{D}_0$
\FOR{$t = 0, \dots, T-1$}
    \STATE Fit probabilistic surrogate to $\mathcal{D}_t$
    \STATE Construct acquisition function $a_t$
    \STATE Select
    $\boldsymbol{x}_{t+1}
      \in \arg\max_{\boldsymbol{x} \in \mathcal{X}}
      a_t(\boldsymbol{x})$
    \STATE Evaluate the merged model
    $\theta(\boldsymbol{x}_{t+1})$
    to obtain $\boldsymbol{y}_{t+1}$
    \STATE Update dataset
    $\mathcal{D}_{t+1}
      = \mathcal{D}_t
      \cup \{(\boldsymbol{x}_{t+1}, \boldsymbol{y}_{t+1})\}$
\ENDFOR
\STATE Return the nondominated solutions among the evaluated configurations
\end{algorithmic}
\endgroup

\end{algorithm}

\section{Experiments and Analysis}
\label{sec:experiments}
We evaluate MOBO-Merge on instruction-following, mathematical-reasoning, and code-generation objectives using two model families. Our experiments address three questions: (i) when does guided multi-objective search improve over random parameter selection under the same evaluation budget, (ii) how does the merge operator affect the quality and coverage of the recovered Pareto set, and (iii) do the same trends hold when moving from two source models and two objectives (AB) to three source models and three objectives (ABC)?

\subsection{Experimental Setup}

\paragraph{Models.}
We evaluate MOBO-Merge on two model families and sizes: Qwen3-4B \citep{qwen3_4b} and Llama-3.1-8B \citep{llama3_8b}. For each family, we use a shared base model together with an instruction-following checkpoint (A), a mathematical-reasoning checkpoint (B), and a third checkpoint (C) used in the code-generation objective. For Qwen3-4B, C is the code-oriented CodeScout checkpoint; for Llama-3.1-8B, C is the continually pretrained Swallow checkpoint, which includes enhanced reasoning and coding capabilities. The primary AB setting merges the instruction and math checkpoints. The ABC setting extends the same framework to a three-model merge by adding C. The full list of checkpoints is provided in Appendix~\ref{app:modellinks}.


\paragraph{Objectives and data splits.}
We measure instruction following with instruction-level strict accuracy on IFEval \citep{ifeval}, mathematical reasoning with exact match under flexible answer extraction on zero-shot chain-of-thought GSM8K \citep{gsm8k}, and code generation with pass@1 on \texttt{humaneval\_instruct} \citep{humaneval}. Each benchmark is deterministically partitioned using a set split seed. The first 30\% of examples form the optimization partition and the complementary 70\% form a held-out partition. Every method and seed uses the same partition. Only configurations on a run's validation Pareto front are re-evaluated on the held-out partition. All evaluations are zero-shot and use deterministic decoding ($\text{temperature}=0$) through the LM Evaluation Harness \citep{eval-harness}.

\paragraph{Merge and evaluation settings.}
Merges are materialized with MergeKit \citep{goddard2024arcee} in FP16 and are deleted after evaluation. We use a maximum model length of 2048 tokens and up to 1024 generated tokens for GSM8K and HumanEval. The evaluation batch size is 128 for Qwen3-4B and 64 for Llama-3.1-8B. Hypervolume is computed with fixed reference points selected separately for each family and objective set: $(0.32,0.64)$ and $(0.32,0.64,0.66)$ for Qwen AB and ABC, and $(0.31,0.48)$ and $(0.23,0.46,0.31)$ for Llama AB and ABC, respectively. Consequently, absolute hypervolume values should only be compared within the same family and objective set.

\paragraph{Search protocol and baseline.}
For each merge operator, we compare sequential NEHVI against uniform random search over the same feasible parameter region. Results are aggregated across seeds 42--51. In the AB experiments, the initial BO design contains 4 points for Linear and 10 points for TIES and Block-Linear 4x. In the ABC experiments, the corresponding initial-design sizes are 6, 14, and 18. These sizes reflect the search dimensions: 1, 4, and 4 in AB, and 2, 6, and 8 in ABC. The three-model weights are constrained to the simplex, with the weight of model C defined by the remaining mass. Each BO run then performs 100 sequential acquisition steps; the random baseline evaluates 100 configurations.



\subsection{Merging Strategies}

We consider four merging strategies. The main comparison includes operators
that admit a consistent parameterization in both the two-model and
three-model settings. Standard SLERP is restricted to two-model interpolation
and has no unique order-independent extension to three-way merging; moreover,
in the AB setting it induces the same one-dimensional search space as Linear
interpolation. We therefore include SLERP only in the Qwen3-4B AB
acquisition-function ablation in Appendix~\ref{app:qnparego}. We
use MergeKit \citep{goddard2024arcee} to implement merging efficiently on
a single GPU (Appendix~\ref{app:compute}).

\paragraph{Linear.} \citet{wortsman2022model} use a weighted average of two models:
\[
\theta(x) = x \theta_0 + (1-x)\theta_1, \qquad x\in [0,1].
\]

\paragraph{SLERP.} \citet{shoemake1985animating} interpolate along the great-circle on the hypersphere:
\[
\theta(x)
= \frac{\sin\!\big((1-x)\Omega\big)}{\sin\Omega}\,\theta_0
+ \frac{\sin\!\big(x\Omega\big)}{\sin\Omega}\,\theta_1, \qquad x\in [0,1],
\]
where \[\Omega = \arccos\!\left(\frac{\langle \theta_0,\theta_1\rangle}{\|\theta_0\|\,\|\theta_1\|}\right).\]

\paragraph{TIES.} \citet{yadav2023ties} mitigate parameter interference by trimming minimally changed parameters, resolving sign conflicts, and merging aligned updates. We use a minor variant that replaces the average of trimmed task vectors with a weighted mean to permit a richer Pareto set. This yields a 4-parameter search space for two-model merging.

\paragraph{Block-wise Linear Merge.}
To study the effect of increasing merge-space dimensionality, we introduce a block-wise variant of linear interpolation. Let the model parameters be structured per-layer tuples
\(\theta=\{\theta^{(1)},\ldots,\theta^{(L)}\}\),
and partition layers into \(k\) disjoint blocks
\(\mathcal{L}=\biguplus_{j=1}^{k}\mathcal{L}_j\).
Define the block selector \(P_{\mathcal{L}_j}\) by
\[
P_{\mathcal{L}_j}(\theta)=\{\tilde{\theta}^{(1)},\ldots,\tilde{\theta}^{(L)}\},
\qquad
\tilde{\theta}^{(\ell)}=
\begin{cases}
\theta^{(\ell)}, & \ell\in\mathcal{L}_j,\\
\mathbf{0}, & \ell\notin\mathcal{L}_j,
\end{cases}
\]
where \(\mathbf{0}\) denotes a tuple of zero tensors matching the shapes of \(\theta^{(\ell)}\).
The \(k\)-parameter block-wise merge is
\[
\theta(\boldsymbol{x})
= \sum_{j=1}^{k} P_{\mathcal{L}_j}\!\Big(x_j\,\theta_0 + (1-x_j)\,\theta_1\Big),
\quad
\boldsymbol{x}=(x_1,\ldots,x_k).
\]
For the main comparison we use four equally sized layer blocks. This gives four independent mixing coefficients in AB and two simplex coordinates per block (eight free parameters) in ABC. Qwen3-4B has 36 transformer layers and Llama-3.1-8B has 32.

\subsection{Bayesian Optimization Loop}
We implement a sequential multi-objective Bayesian optimization loop in BoTorch \citep{balandat2020botorch}. The loop begins with a Sobol design of $n_{\text{init}}$ initial configurations, augmented with endpoint settings that recover each source model. After each evaluation, we fit independent GP surrogates: one \texttt{SingleTaskGP} per objective, combined as a \texttt{ModelListGP}, with input normalization to $[0,1]^d$ and outcome standardization; hyperparameters are learned by maximizing the summed marginal log-likelihood. New candidates are proposed by maximizing Noisy Expected Hypervolume Improvement (NEHVI) \citep{daulton2021parallel} using a Sobol QMC sampler with 128 samples and multi-start acquisition optimization. For the three-model setting, acquisition optimization also respects the simplex constraints on the merge weights. We ablate qNParEGO in Appendix~\ref{app:qnparego} and find that neither acquisition function is uniformly more effective across merge operators.

\subsection{Performance Across Operators}
\begin{figure}[t]
    \centering

    \begin{subfigure}[t]{0.42\linewidth}
        \centering
        \includegraphics[width=\linewidth]{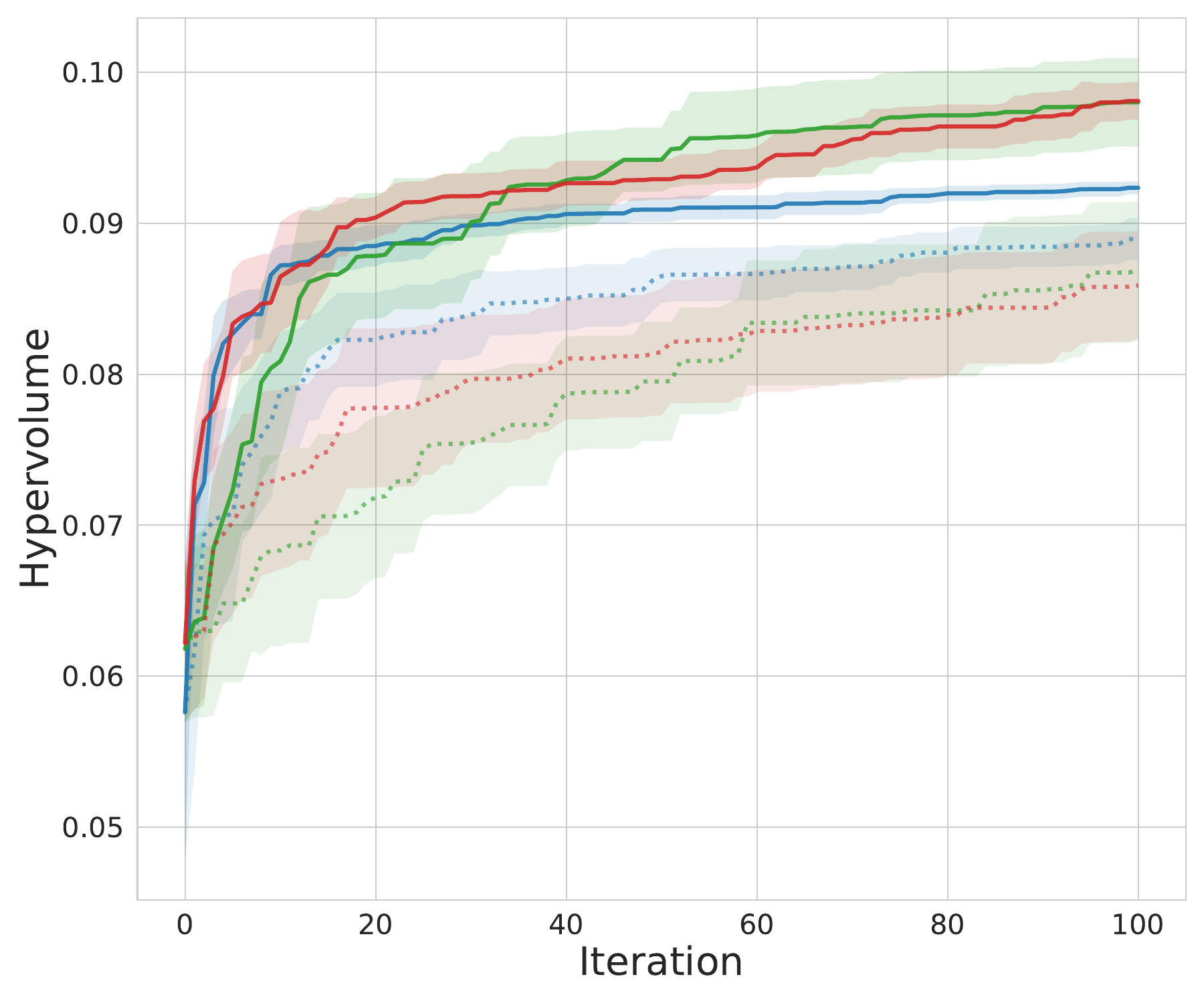}
        \caption{Qwen3 4B models}
    \end{subfigure}
    \hfill
    \begin{subfigure}[t]{0.42\linewidth}
        \centering
        \includegraphics[width=\linewidth]{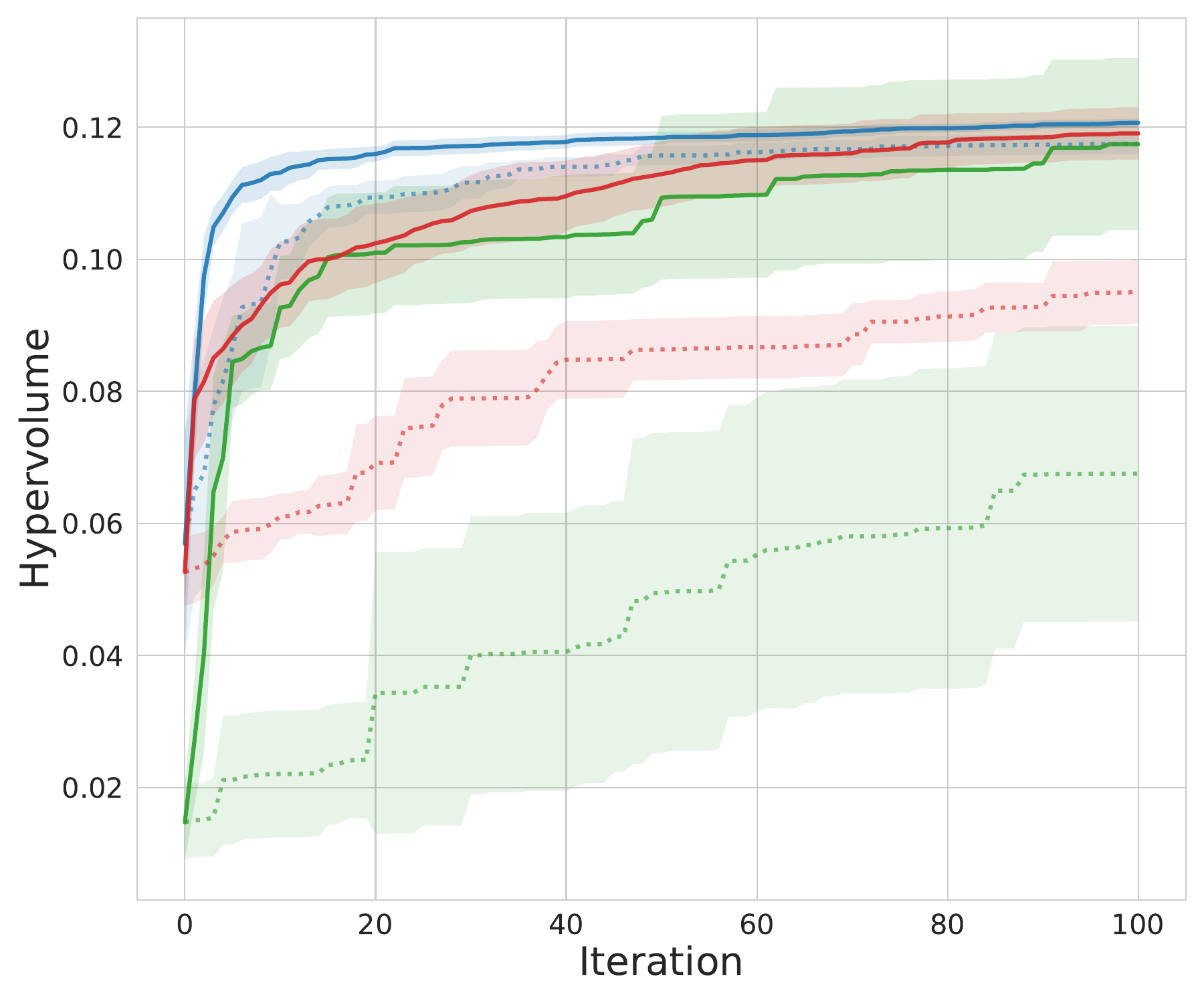}
        \caption{Llama3 8B models}
    \end{subfigure}
    \hfill
    \begin{subfigure}[t]{0.14\linewidth}
        \centering
        \includegraphics[width=\linewidth]{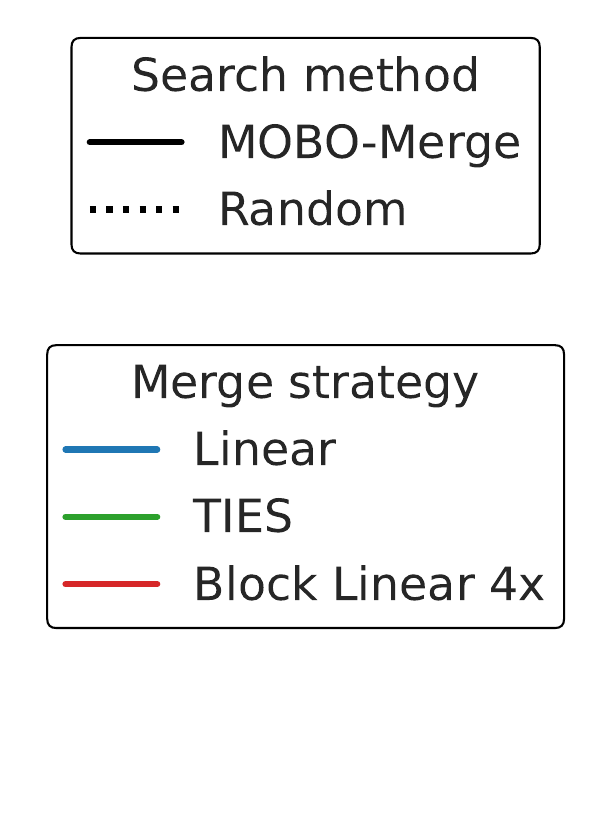}
    \end{subfigure}

    \caption{\textbf{Validation hypervolume in the two-model AB setting.} Running dominated hypervolume for Linear, TIES, and Block-Linear 4x on instruction following and mathematical reasoning. Solid curves show MOBO-Merge with NEHVI and dotted curves show random search; curves show the mean across ten seeds, and shaded regions indicate the mean $\pm$ one standard error. The Linear gap is small, while MOBO-Merge produces clearer gains for the higher-dimensional TIES and Block-Linear search spaces, particularly for Llama-3.1-8B. 
    }
    \label{fig:block}
\end{figure}

\begin{figure}[t]
    \centering
    \begin{subfigure}[t]{0.42\linewidth}
        \centering
        \includegraphics[width=\linewidth]{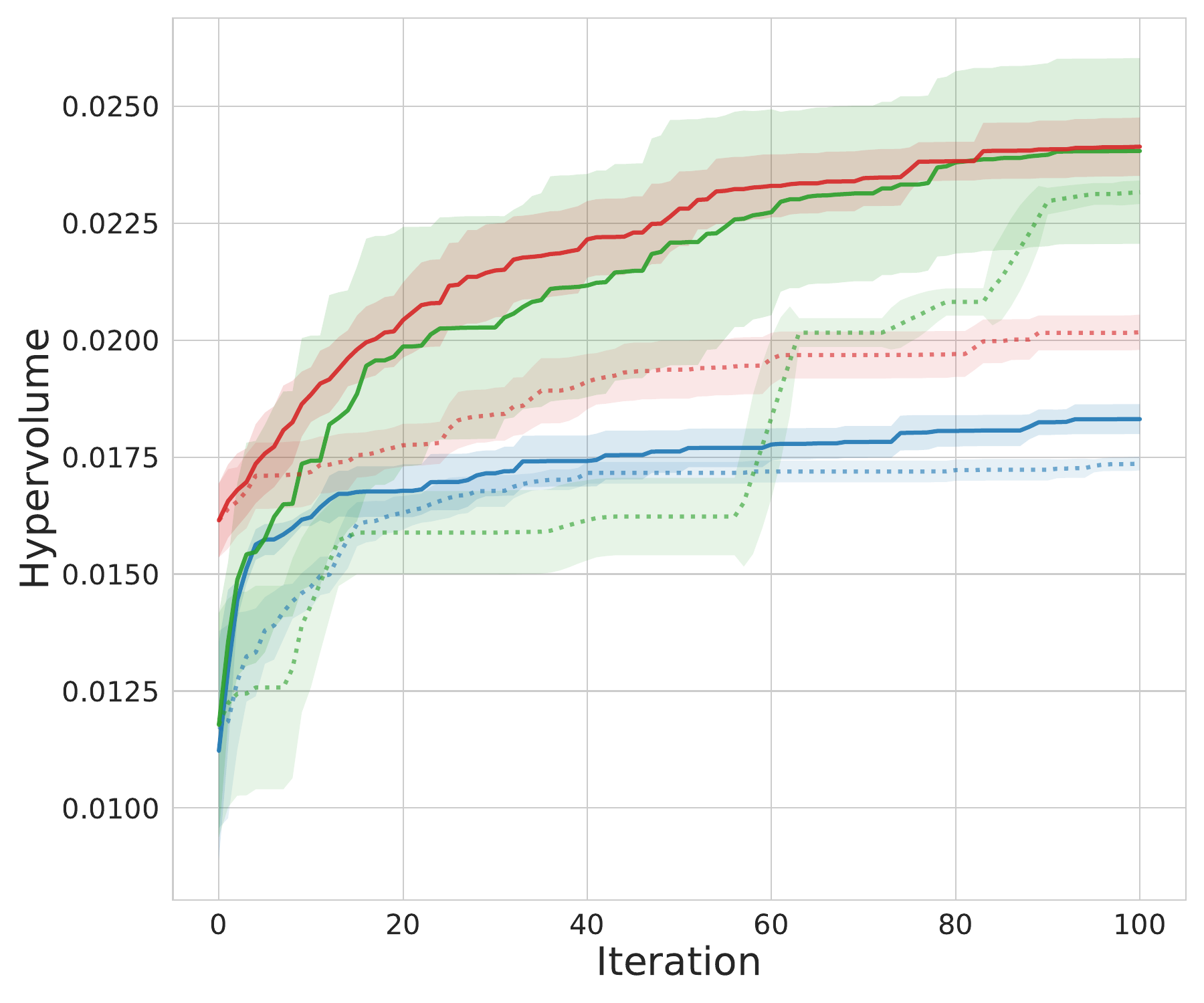}
        \caption{Qwen3 4B models}
    \end{subfigure}
    \hfill
    \begin{subfigure}[t]{0.42\linewidth}
        \centering
        \includegraphics[width=\linewidth]{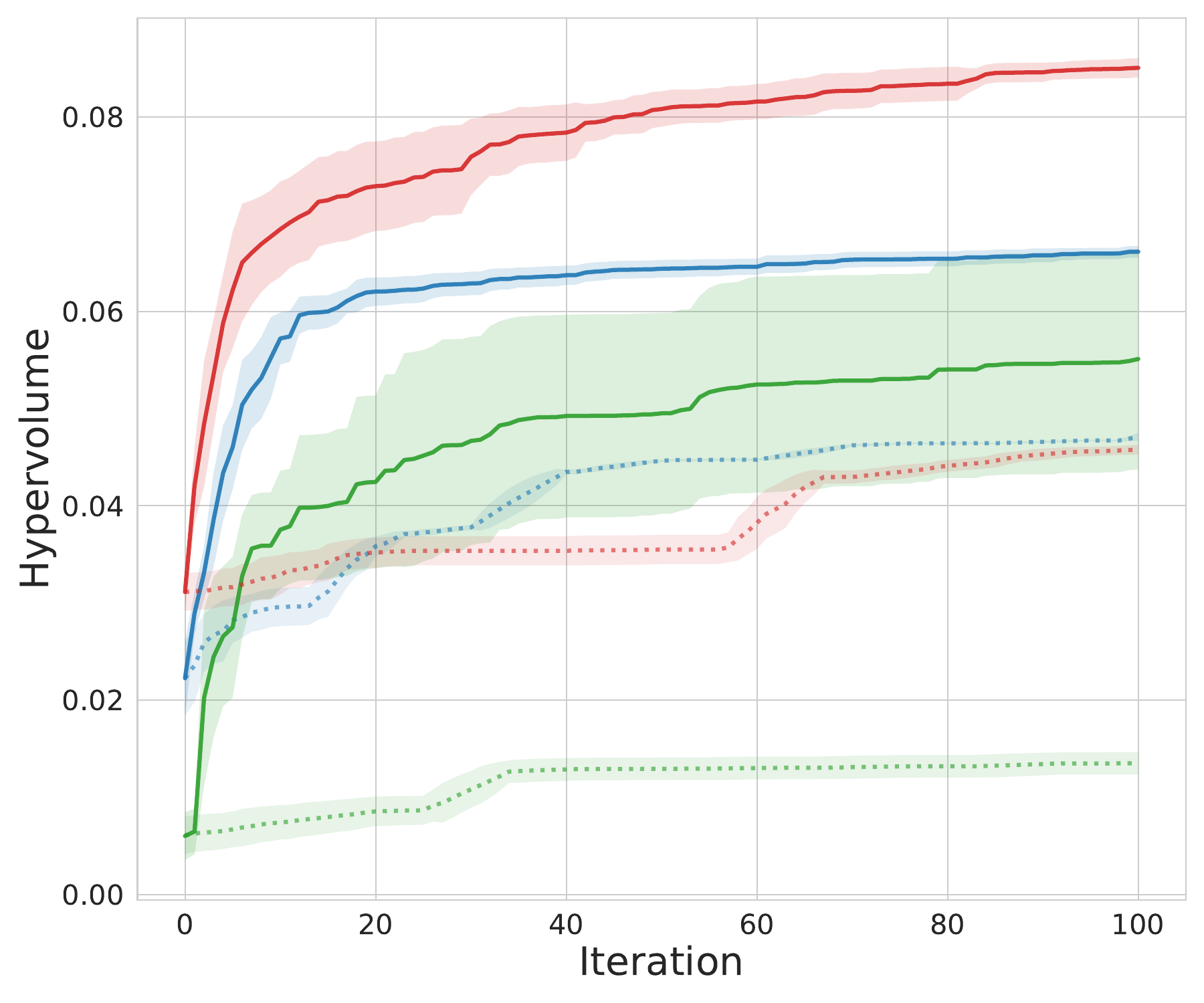}
        \caption{Llama3 8B models}
    \end{subfigure}
    \hfill
    \begin{subfigure}[t]{0.14\linewidth}
        \centering
        \includegraphics[width=\linewidth]{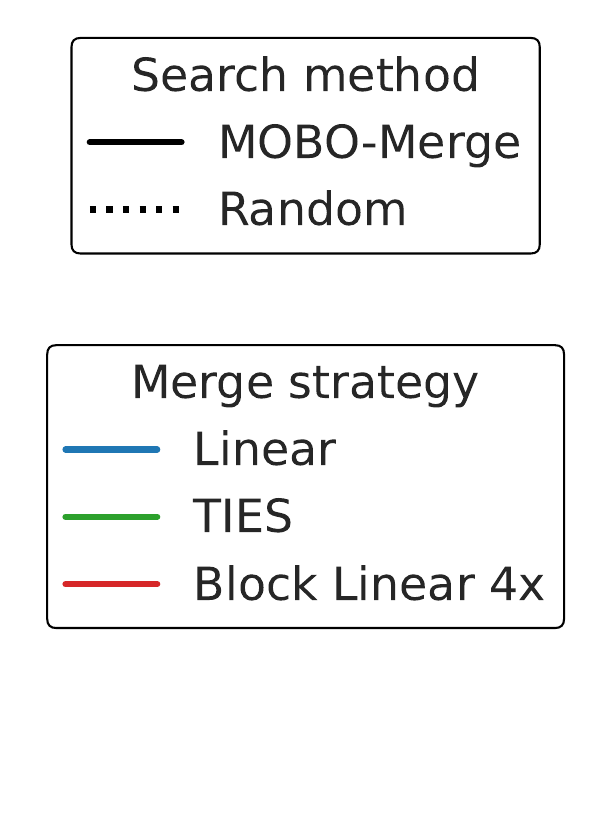}
    \end{subfigure}
    \caption{\textbf{Validation hypervolume in the three-model ABC setting.} Running dominated hypervolume for Linear, TIES, and Block-Linear 4x on instruction following, mathematical reasoning, and code generation. Solid curves show MOBO-Merge with NEHVI and dotted curves show random search; curves show the mean across ten seeds, and shaded regions indicate the mean $\pm$ one standard error. The advantage of guided search is larger than in the AB setting, most notably for Llama-3.1-8B, where Block-Linear 4x attains the highest final hypervolume.
    }
    \label{fig:ties-block}
\end{figure}


\paragraph{Two-model results.}
Figure~\ref{fig:block} shows that the benefit of MOBO-Merge depends strongly on the merge space. Linear interpolation is one-dimensional, and random search is already competitive: on the held-out Qwen AB split, random search is slightly higher than MOBO-Merge (0.0664 versus 0.0659), while MOBO-Merge has a small advantage for Llama AB (0.1036 versus 0.1017). The advantage is clearer for the four-dimensional operators. Relative to random search, MOBO-Merge raises held-out hypervolume from 0.0739 to 0.0847 for Qwen TIES and from 0.0677 to 0.0737 for Qwen Block-Linear 4x. For Llama, the corresponding improvements are from 0.0764 to 0.1145 and from 0.0880 to 0.1057. Thus, guided search is most useful when the operator exposes a sufficiently rich parameter space; it is less consequential for a single global interpolation coefficient.

\paragraph{Three-model results.}
The advantage widens in the ABC setting (Figure~\ref{fig:ties-block}). On Qwen3-4B, MOBO-Merge improves held-out hypervolume over random search for all three operators: 0.0131 versus 0.0129 for Linear, 0.0184 versus 0.0145 for TIES, and 0.0161 versus 0.0144 for Block-Linear 4x. On Llama-3.1-8B, the gains are larger: 0.0430 versus 0.0367 for Linear, 0.0429 versus 0.0055 for TIES, and 0.0634 versus 0.0289 for Block-Linear 4x. The best operator is not universal. TIES has the highest held-out hypervolume in both Qwen settings and in Llama AB, whereas Block-Linear 4x is strongest in Llama ABC.

\paragraph{Held-out generalization.}
Across the 12 family, objective-set, and operator comparisons in Table~\ref{table:hv-test}, MOBO-Merge obtains the higher mean held-out hypervolume in 11. The only reversal is Qwen AB with Linear, where random search is higher by 0.0005 (0.7\% relative). The benefit is small in a one-dimensional space but substantial for several higher-dimensional and three-objective searches. The held-out results also show that TIES is not uniformly robust to unguided parameter selection; its random-search variance and performance degradation are especially pronounced for Llama.

\begin{table}[t]
\caption{\textbf{Mean hypervolume on the held-out partition.} For each run, the validation Pareto set is re-evaluated on the complementary partition and its held-out hypervolume is computed. Cells average over 10 completed runs. Reference points differ by family and objective set, so values should be compared only within rows, not across rows. Bold denotes the best value in each row.}
\label{table:hv-test}
\begin{center}
\begin{tabular}{cc|cccccc}
\multicolumn{1}{l}{} &
  \multicolumn{1}{l|}{} &
  \multicolumn{2}{c|}{Linear} &
  \multicolumn{2}{c|}{TIES} &
  \multicolumn{2}{c}{Block-Linear 4x} \\ \cline{3-8} 
\multicolumn{1}{l}{} &
  \multicolumn{1}{l|}{} &
  \multicolumn{1}{c|}{MOBO-Merge} &
  \multicolumn{1}{c|}{Random} &
  \multicolumn{1}{c|}{MOBO-Merge} &
  \multicolumn{1}{c|}{Random} &
  \multicolumn{1}{c|}{MOBO-Merge} &
  \multicolumn{1}{c}{Random} \\ \hline
\multicolumn{1}{c|}{\multirow{2}{*}{4B}} & AB  &  0.0659
 & 0.0664 & \textbf{0.0847} & 0.0739 & 0.07368 & 0.0677 \\ \cline{2-2}
\multicolumn{1}{c|}{}                    & ABC & 0.0131 & 0.0129 & \textbf{0.0184} & 0.0145 & 0.0161 & 0.0144 \\ \cline{1-2}
\multicolumn{1}{c|}{\multirow{2}{*}{8B}} & AB  & 0.1036 & 0.1017 & \textbf{0.1145} & 0.0764 & 0.1057 & 0.0879 \\ \cline{2-2}
\multicolumn{1}{c|}{}                    & ABC & 0.0430 & 0.0367 & 0.0429 & 0.0055 & \textbf{0.0634} & 0.0289
\end{tabular}
\end{center}
\end{table}

\subsection{Pareto Front Recovery}
\begin{figure}[h]
  \vskip 0.2in
  \centering
  \begin{subfigure}[t]{0.43\linewidth}
        \centering
        \includegraphics[width=\linewidth]{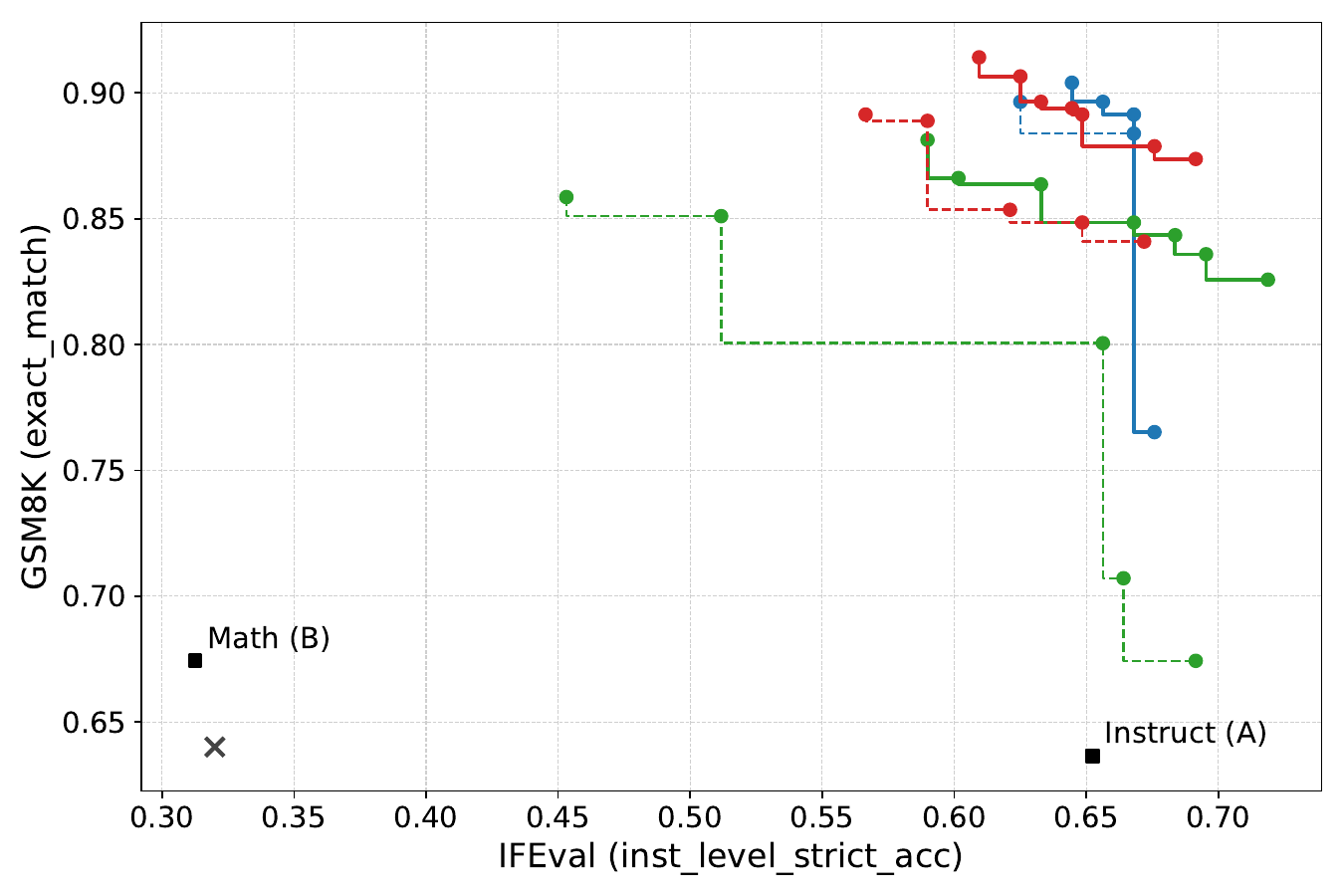}
        \caption{Qwen3-4B}
    \end{subfigure}
    \hfill
    \begin{subfigure}[t]{0.43\linewidth}
        \centering
        \includegraphics[width=\linewidth]{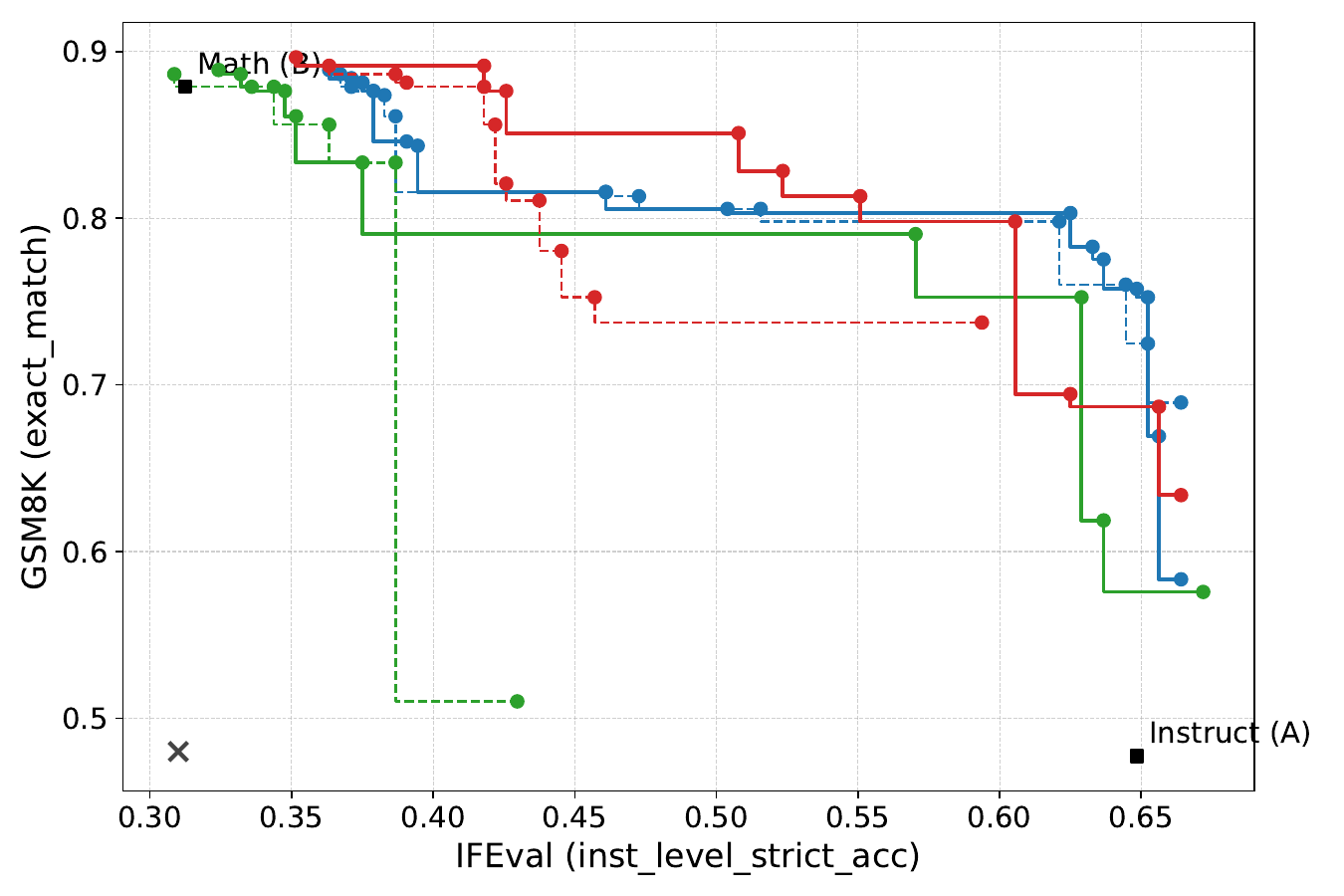}
        \caption{Llama3.1-8B}
    \end{subfigure}
    \hfill
    \begin{subfigure}[t]{0.12\linewidth}
        \centering
        \includegraphics[width=\linewidth]{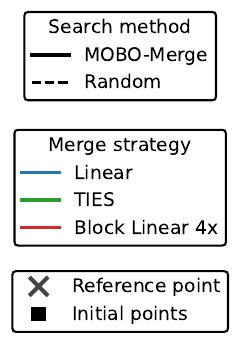}
    \end{subfigure}
    \caption{\textbf{Representative validation Pareto fronts for instruction following and mathematical reasoning.} Solid lines denote MOBO-Merge, dashed lines denote random search, and black squares mark the two source models. The Qwen front contains merges that improve both annotated source scores simultaneously, whereas the Llama front retains a clear capability trade-off.}
    \label{fig:paretoAB}
    \end{figure}

Figure~\ref{fig:paretoAB} shows representative validation Pareto fronts for Qwen seed 44 and Llama seed 42. The fronts illustrate why a single scalar summary is insufficient: different operators cover different regions of the instruction-math trade-off. In the displayed Qwen run, all three MOBO operators discover configurations whose IFEval and GSM8K scores both exceed the two annotated source checkpoints, although Block-Linear 4x has the largest hypervolume and TIES does not dominate the Linear front. In the displayed Llama run, no configuration exceeds the coordinate-wise maximum of both sources; the problem remains a genuine trade-off, and Block-Linear 4x covers more of the high-value region than Linear or TIES. These single-seed plots are illustrative; the across-seed conclusions are given by Figures~\ref{fig:block} and \ref{fig:ties-block} and Table~\ref{table:hv-test}.


\subsection{Discussion}
Our experiments support three main takeaways. First, model merging induces complex trade-off landscapes in which naive parameter search can waste evaluations, while MOBO-Merge concentrates evaluations in regions that expand the Pareto front. The benefit nevertheless depends on the complexity of the search space: random search remains competitive for one-dimensional Linear interpolation, whereas MOBO-Merge is more consistently advantageous for TIES, block-wise merging, and the three-model setting. Second, increasing the expressivity of the merge space through block-wise parameterizations can yield stronger trade-offs, but makes sample-efficient optimization increasingly important. In particular, Block-Linear 4x produces the best Llama ABC result and more than doubles the held-out hypervolume of random search in that setting while remaining simple and interpretable. Third, the merge operator and search strategy must be considered jointly. TIES gives the strongest held-out result in three of four family-setting combinations under MOBO-Merge, but is not uniformly robust to parameter choice and performs poorly under random selection in several Llama experiments. Validation improvements generally transfer to the held-out partition, although the small reversal for Qwen AB with Linear shows that the advantage is not universal. Overall, MOBO-Merge provides a practical framework for controllable model composition under limited evaluation budgets and is most useful for efficiently exploring expressive merge spaces.





\section{Related Work}

\subsection{Model Merging and Weight-Space Composition}
Model merging\footnote{Throughout this paper, we use \emph{model merging} to refer to training-free composition via weight-space operations. We use the broader term \emph{model fusion} for approaches that may involve additional training or non–weight-space combination.} has emerged as a practical alternative to additional pretraining or joint multi-task finetuning, particularly for large language models  \citep{yang2024model}. Instead of training a single monolithic model, practitioners train specialized models and combine them post hoc in weight space.
Early and influential work shows that simple averaging or interpolation of fine-tuned models can improve robustness and accuracy without increasing inference cost, as in model soups~\citep{wortsman2022model}.
Task arithmetic further demonstrates that linear combinations of task-specific parameter updates can compose behaviors such as domain adaptation and style control~\citep{ilharco2022editing}. Finally, recent theoretical work suggests that model fusion and merging can admit non-vacuous generalization guarantees in certain regimes, supporting the view that merging can be both practical and principled~\citep{kim2025model}.

More recent work emphasizes mitigating interference between task updates and introducing structure into the merge space.
TIES-Merging explicitly resolves destructive parameter conflicts to better preserve multiple capabilities~\citep{yadav2023ties}, while DARE rescales and prunes parameter deltas to improve compatibility between homologous language models~\citep{yu2024language}.
Other methods explore structured merge parameterizations, such as  block-wise coefficients, to increase expressivity while keeping optimization tractable.
For example, Parameter Competition Balancing (PCB) balances competing parameter contributions during merging~\citep{du2024parameter}, and KAFA-Merge partitions layers into groups and applies block-wise linear interpolation with optimized mixing coefficients~\citep{yuce2025kafa}.
Tooling efforts such as MergeKit further systematize these approaches and make large-scale merging practical~\citep{goddard2024arcee}.

While these methods improve the \emph{merge operator} and its parameterization, they also expose a recurring challenge: even with a well-designed operator, selecting merge coefficients is an expensive, noisy, and often nontrivial optimization problem, especially when multiple capabilities must be preserved simultaneously.

\subsection{Bayesian Optimization}

Bayesian optimization (BO) is a widely used framework for optimizing expensive black-box functions under limited evaluation budgets \citep{frazier2018tutorial,garnett2023bayesian}. Originally popularized for hyperparameter optimization in machine learning \citep{snoek2012practical}, BO has since become a standard tool for problems characterized by costly, noisy, and gradient-free evaluations. More recently, its scope has expanded beyond model tuning to scientific discovery and experimental design in domains such as materials science, chemistry, and engineering, where each evaluation may correspond to a real-world experiment or simulation \citep{cosenza2022multi,griffiths2020constrained,mathern2021multi}. Advances in probabilistic surrogate modeling and acquisition function design have further extended BO to structured search spaces, noisy objectives, and settings with multiple competing objectives \citep{daulton2021parallel,eriksson2021high,wu2020practical}.

\subsection{Bayesian Optimization for Model Merging and Fusion}
Recent work has explored Bayesian optimization (BO) as a principled approach for selecting merge weights when model evaluation is expensive, noisy, and gradient-free. These efforts motivate a black-box optimization view of model merging and fusion, and span a range of settings with different structural assumptions and optimization objectives.

Several lines of work apply BO to model merging under a \emph{single-objective} formulation. This includes approaches that use BO to merge checkpoints sampled from a single training trajectory or continual learning process, typically optimizing a single downstream objective under assumptions about the structure of the optimization path \citep{liu2024checkpoint,li2025became}. Related work applies BO to post-hoc model combination over structured merge spaces, such as optimizing group-wise linear interpolation weights (sometimes followed by local refinement), again with the goal of identifying a single high-performing merged model under a scalar evaluation metric \citep{yuce2025kafa}. While effective in their respective regimes, these approaches focus on selecting a single merged model and do not explicitly characterize trade-offs between heterogeneous capabilities.

More closely related work adopts a \emph{multi-objective} perspective on
model composition. In model merging, existing approaches optimize merge
parameters to recover solutions representing different trade-offs between
objectives \citep{li2025s,chen2025bambo}. These methods establish the value of
Pareto-based search, but generally couple the optimization procedure to a
particular merge operator, parameterization, or target trade-off, such as
balancing model ability against computational efficiency. Related work on
model fusion applies multi-objective Bayesian optimization to averaging
coefficients for checkpoints obtained from a single fine-tuning process,
typically balancing loss and task-specific performance to improve
generalization or mitigate overfitting \citep{jang2024model}.

In contrast to prior work, we study multi-objective Bayesian optimization as
a common search layer across heterogeneous model-merging operators. We use a
fixed optimization and evaluation protocol while varying the merge
parameterization, model family, and number of source models and capability
objectives. This controlled design enables direct comparison of global
interpolation, interference-aware merging, and block-wise parameterizations,
and reveals how the value of guided search depends on the structure and
dimensionality of the merge space. We further re-evaluate configurations from
the optimization-partition Pareto set on complementary held-out benchmark
partitions to assess whether the discovered capability trade-offs transfer
beyond the data used to guide the search.

\section{Limitations}

MOBO-Merge relies on repeated evaluations of merged models on downstream benchmarks, which can be computationally expensive. Although the method is substantially more sample-efficient than naive search strategies, evaluation cost remains a practical bottleneck, especially when objectives require long-running inference or exhibit high variance. This limits the total number of merge configurations that can be explored in practice.

Our framework focuses on weight-space merging of models that share a common architecture and parameterization. While this setting covers many practical merging scenarios for large language models, extending MOBO-Merge to heterogeneous models or non–weight-based composition methods remains an open challenge. Moreover, the quality of the recovered Pareto front depends on the choice of evaluation benchmarks and objectives, which may not fully capture all aspects of model behavior or generalization.

Finally, while we demonstrate strong performance in low- to moderate-dimensional merge spaces with a small number of objectives, scaling MOBO-Merge to settings with higher-dimensional merge spaces and many competing objectives may require more expressive surrogate models and improved acquisition strategies.

\section{Conclusion}

We introduced \textbf{MOBO-Merge}, a framework that formulates model merging as an expensive black-box multi-objective optimization problem and applies multi-objective Bayesian optimization to efficiently discover high-quality merges under limited evaluation budgets. By explicitly targeting Pareto-optimal solutions, MOBO-Merge exposes controllable trade-offs between competing model capabilities and replaces ad hoc merge-weight tuning with a principled optimization-based approach.

Our experiments on Qwen3-4B and Llama-3.1-8B show that MOBO-Merge reliably recovers well-structured Pareto fronts when merging specialized large language models, while requiring fewer evaluations than random search. 
Across held-out comparisons, MOBO-Merge achieves higher mean hypervolume than random search in 11 of 12 settings, with the gains most pronounced in several higher-dimensional TIES, block-wise, and three-model searches.
Moreover, the framework is merge-operator agnostic and extends from two-model to three-model merging, demonstrating that standard multi-objective Bayesian optimization provides a practical and general tool for model composition without additional pretraining.

Looking ahead, promising directions include adaptive merge parameterizations that learn which layers or components require independent weights, more efficient evaluation strategies such as early stopping or multi-fidelity signals, and extensions to higher-dimensional merge spaces or larger numbers of objectives.





\subsubsection*{Broader Impact Statement}
This paper presents work whose goal is to improve the practice of model merging in machine learning by providing a systematic and compute-efficient way to explore trade-offs between model capabilities. The proposed framework applies existing multi-objective optimization techniques to the problem of merging pretrained models and does not introduce new optimization methods or new model capabilities beyond those present in the source checkpoints. By enabling more flexible reuse of existing models, this work may help reduce the need for additional large-scale training. We do not anticipate specific societal impacts arising uniquely from this work beyond those commonly associated with the deployment of large language models.



\bibliography{mainbib}

\begin{thebibliography}{34}
\providecommand{\natexlab}[1]{#1}
\providecommand{\url}[1]{\texttt{#1}}
\expandafter\ifx\csname urlstyle\endcsname\relax
  \providecommand{\doi}[1]{doi: #1}\else
  \providecommand{\doi}{doi: \begingroup \urlstyle{rm}\Url}\fi

\bibitem[Balandat et~al.(2020)Balandat, Karrer, Jiang, Daulton, Letham, Wilson,
  and Bakshy]{balandat2020botorch}
Maximilian Balandat, Brian Karrer, Daniel Jiang, Samuel Daulton, Ben Letham,
  Andrew~G Wilson, and Eytan Bakshy.
\newblock Botorch: A framework for efficient monte-carlo bayesian optimization.
\newblock \emph{Advances in neural information processing systems},
  33:\penalty0 21524--21538, 2020.

\bibitem[Chen et~al.(2025)Chen, Luo, Zhu, Hu, and Xi]{chen2025bambo}
Kesheng Chen, Wenjian Luo, Zhenqian Zhu, Yamin Hu, and Yiya Xi.
\newblock Bambo: Construct ability and efficiency llm pareto set via bayesian
  adaptive multi-objective block-wise optimization.
\newblock \emph{arXiv preprint arXiv:2512.09972}, 2025.

\bibitem[Chen et~al.(2021)Chen, Tworek, Jun, Yuan, de~Oliveira~Pinto, Kaplan,
  Edwards, Burda, Joseph, Brockman, Ray, Puri, Krueger, Petrov, Khlaaf, Sastry,
  Mishkin, Chan, Gray, Ryder, Pavlov, Power, Kaiser, Bavarian, Winter, Tillet,
  Such, Cummings, Plappert, Chantzis, Barnes, Herbert-Voss, Guss, Nichol,
  Paino, Tezak, Tang, Babuschkin, Balaji, Jain, Saunders, Hesse, Carr, Leike,
  Achiam, Misra, Morikawa, Radford, Knight, Brundage, Murati, Mayer, Welinder,
  McGrew, Amodei, McCandlish, Sutskever, and Zaremba]{humaneval}
Mark Chen, Jerry Tworek, Heewoo Jun, Qiming Yuan, Henrique~Ponde
  de~Oliveira~Pinto, Jared Kaplan, Harri Edwards, Yuri Burda, Nicholas Joseph,
  Greg Brockman, Alex Ray, Raul Puri, Gretchen Krueger, Michael Petrov, Heidy
  Khlaaf, Girish Sastry, Pamela Mishkin, Brooke Chan, Scott Gray, Nick Ryder,
  Mikhail Pavlov, Alethea Power, Lukasz Kaiser, Mohammad Bavarian, Clemens
  Winter, Philippe Tillet, Felipe~Petroski Such, Dave Cummings, Matthias
  Plappert, Fotios Chantzis, Elizabeth Barnes, Ariel Herbert-Voss,
  William~Hebgen Guss, Alex Nichol, Alex Paino, Nikolas Tezak, Jie Tang, Igor
  Babuschkin, Suchir Balaji, Shantanu Jain, William Saunders, Christopher
  Hesse, Andrew~N. Carr, Jan Leike, Josh Achiam, Vedant Misra, Evan Morikawa,
  Alec Radford, Matthew Knight, Miles Brundage, Mira Murati, Katie Mayer, Peter
  Welinder, Bob McGrew, Dario Amodei, Sam McCandlish, Ilya Sutskever, and
  Wojciech Zaremba.
\newblock Evaluating large language models trained on code.
\newblock 2021.

\bibitem[Cobbe et~al.(2021)Cobbe, Kosaraju, Bavarian, Chen, Jun, Kaiser,
  Plappert, Tworek, Hilton, Nakano, et~al.]{gsm8k}
Karl Cobbe, Vineet Kosaraju, Mohammad Bavarian, Mark Chen, Heewoo Jun, Lukasz
  Kaiser, Matthias Plappert, Jerry Tworek, Jacob Hilton, Reiichiro Nakano,
  et~al.
\newblock Training verifiers to solve math word problems.
\newblock \emph{arXiv preprint arXiv:2110.14168}, 2021.

\bibitem[Cosenza et~al.(2022)Cosenza, Astudillo, Frazier, Baar, and
  Block]{cosenza2022multi}
Zachary Cosenza, Raul Astudillo, Peter~I Frazier, Keith Baar, and David~E
  Block.
\newblock Multi-information source bayesian optimization of culture media for
  cellular agriculture.
\newblock \emph{Biotechnology and bioengineering}, 119\penalty0 (9):\penalty0
  2447--2458, 2022.

\bibitem[Daulton et~al.(2020)Daulton, Balandat, and
  Bakshy]{daulton2020differentiable}
Samuel Daulton, Maximilian Balandat, and Eytan Bakshy.
\newblock Differentiable expected hypervolume improvement for parallel
  multi-objective bayesian optimization.
\newblock \emph{Advances in neural information processing systems},
  33:\penalty0 9851--9864, 2020.

\bibitem[Daulton et~al.(2021)Daulton, Balandat, and
  Bakshy]{daulton2021parallel}
Samuel Daulton, Maximilian Balandat, and Eytan Bakshy.
\newblock Parallel bayesian optimization of multiple noisy objectives with
  expected hypervolume improvement.
\newblock \emph{Advances in neural information processing systems},
  34:\penalty0 2187--2200, 2021.

\bibitem[Du et~al.(2024)Du, Lee, Li, Jiang, Guo, Yu, Liu, Goh, Tang, He,
  et~al.]{du2024parameter}
Guodong Du, Junlin Lee, Jing Li, Runhua Jiang, Yifei Guo, Shuyang Yu, Hanting
  Liu, Sim~K Goh, Ho-Kin Tang, Daojing He, et~al.
\newblock Parameter competition balancing for model merging.
\newblock \emph{Advances in Neural Information Processing Systems},
  37:\penalty0 84746--84776, 2024.

\bibitem[Eriksson \& Jankowiak(2021)Eriksson and Jankowiak]{eriksson2021high}
David Eriksson and Martin Jankowiak.
\newblock High-dimensional bayesian optimization with sparse axis-aligned
  subspaces.
\newblock In \emph{Uncertainty in Artificial Intelligence}, pp.\  493--503.
  PMLR, 2021.

\bibitem[Frazier(2018)]{frazier2018tutorial}
Peter~I Frazier.
\newblock A tutorial on bayesian optimization.
\newblock \emph{arXiv preprint arXiv:1807.02811}, 2018.

\bibitem[Gao et~al.(2024)Gao, Tow, Abbasi, Biderman, Black, DiPofi, Foster,
  Golding, Hsu, Le~Noac'h, Li, McDonell, Muennighoff, Ociepa, Phang, Reynolds,
  Schoelkopf, Skowron, Sutawika, Tang, Thite, Wang, Wang, and
  Zou]{eval-harness}
Leo Gao, Jonathan Tow, Baber Abbasi, Stella Biderman, Sid Black, Anthony
  DiPofi, Charles Foster, Laurence Golding, Jeffrey Hsu, Alain Le~Noac'h,
  Haonan Li, Kyle McDonell, Niklas Muennighoff, Chris Ociepa, Jason Phang,
  Laria Reynolds, Hailey Schoelkopf, Aviya Skowron, Lintang Sutawika, Eric
  Tang, Anish Thite, Ben Wang, Kevin Wang, and Andy Zou.
\newblock The language model evaluation harness, 07 2024.
\newblock URL \url{https://zenodo.org/records/12608602}.

\bibitem[Garnett(2023)]{garnett2023bayesian}
Roman Garnett.
\newblock \emph{Bayesian optimization}.
\newblock Cambridge University Press, 2023.

\bibitem[Goddard et~al.(2024)Goddard, Siriwardhana, Ehghaghi, Meyers,
  Karpukhin, Benedict, McQuade, and Solawetz]{goddard2024arcee}
Charles Goddard, Shamane Siriwardhana, Malikeh Ehghaghi, Luke Meyers, Vlad
  Karpukhin, Brian Benedict, Mark McQuade, and Jacob Solawetz.
\newblock Arcee's mergekit: A toolkit for merging large language models.
\newblock \emph{arXiv preprint arXiv:2403.13257}, 2024.

\bibitem[Grattafiori et~al.(2024)Grattafiori, Dubey, Jauhri, Pandey, Kadian,
  Al-Dahle, Letman, Mathur, Schelten, Vaughan, Yang, Fan, Goyal, Hartshorn,
  Yang, Mitra, Sravankumar, Korenev, Hinsvark, Rao, Zhang, Rodriguez,
  Gregerson, Spataru, Roziere, Biron, Tang, Chern, Caucheteux, Nayak, Bi,
  Marra, McConnell, Keller, Touret, Wu, Wong, Ferrer, Nikolaidis, Allonsius,
  Song, Pintz, Livshits, Wyatt, Esiobu, Choudhary, Mahajan, Garcia-Olano,
  Perino, Hupkes, Lakomkin, AlBadawy, Lobanova, Dinan, Smith, Radenovic,
  Guzmán, Zhang, Synnaeve, Lee, Anderson, Thattai, Nail, Mialon, Pang,
  Cucurell, Nguyen, Korevaar, Xu, Touvron, Zarov, Ibarra, Kloumann, Misra,
  Evtimov, Zhang, Copet, Lee, Geffert, Vranes, Park, Mahadeokar, Shah, van~der
  Linde, Billock, Hong, Lee, Fu, Chi, Huang, Liu, Wang, Yu, Bitton, Spisak,
  Park, Rocca, Johnstun, Saxe, Jia, Alwala, Prasad, Upasani, Plawiak, Li,
  Heafield, Stone, El-Arini, Iyer, Malik, Chiu, Bhalla, Lakhotia,
  Rantala-Yeary, van~der Maaten, Chen, Tan, Jenkins, Martin, Madaan, Malo,
  Blecher, Landzaat, de~Oliveira, Muzzi, Pasupuleti, Singh, Paluri, Kardas,
  Tsimpoukelli, Oldham, Rita, Pavlova, Kambadur, Lewis, Si, Singh, Hassan,
  Goyal, Torabi, Bashlykov, Bogoychev, Chatterji, Zhang, Duchenne, Çelebi,
  Alrassy, Zhang, Li, Vasic, Weng, Bhargava, Dubal, Krishnan, Koura, Xu, He,
  Dong, Srinivasan, Ganapathy, Calderer, Cabral, Stojnic, Raileanu, Maheswari,
  Girdhar, Patel, Sauvestre, Polidoro, Sumbaly, Taylor, Silva, Hou, Wang,
  Hosseini, Chennabasappa, Singh, Bell, Kim, Edunov, Nie, Narang, Raparthy,
  Shen, Wan, Bhosale, Zhang, Vandenhende, Batra, Whitman, Sootla, Collot,
  Gururangan, Borodinsky, Herman, Fowler, Sheasha, Georgiou, Scialom,
  Speckbacher, Mihaylov, Xiao, Karn, Goswami, Gupta, Ramanathan, Kerkez,
  Gonguet, Do, Vogeti, Albiero, Petrovic, Chu, Xiong, Fu, Meers, Martinet,
  Wang, Wang, Tan, Xia, Xie, Jia, Wang, Goldschlag, Gaur, Babaei, Wen, Song,
  Zhang, Li, Mao, Coudert, Yan, Chen, Papakipos, Singh, Srivastava, Jain,
  Kelsey, Shajnfeld, Gangidi, Victoria, Goldstand, Menon, Sharma, Boesenberg,
  Baevski, Feinstein, Kallet, Sangani, Teo, Yunus, Lupu, Alvarado, Caples, Gu,
  Ho, Poulton, Ryan, Ramchandani, Dong, Franco, Goyal, Saraf, Chowdhury,
  Gabriel, Bharambe, Eisenman, Yazdan, James, Maurer, Leonhardi, Huang, Loyd,
  Paola, Paranjape, Liu, Wu, Ni, Hancock, Wasti, Spence, Stojkovic, Gamido,
  Montalvo, Parker, Burton, Mejia, Liu, Wang, Kim, Zhou, Hu, Chu, Cai, Tindal,
  Feichtenhofer, Gao, Civin, Beaty, Kreymer, Li, Adkins, Xu, Testuggine, David,
  Parikh, Liskovich, Foss, Wang, Le, Holland, Dowling, Jamil, Montgomery,
  Presani, Hahn, Wood, Le, Brinkman, Arcaute, Dunbar, Smothers, Sun, Kreuk,
  Tian, Kokkinos, Ozgenel, Caggioni, Kanayet, Seide, Florez, Schwarz, Badeer,
  Swee, Halpern, Herman, Sizov, Guangyi, Zhang, Lakshminarayanan, Inan,
  Shojanazeri, Zou, Wang, Zha, Habeeb, Rudolph, Suk, Aspegren, Goldman, Zhan,
  Damlaj, Molybog, Tufanov, Leontiadis, Veliche, Gat, Weissman, Geboski, Kohli,
  Lam, Asher, Gaya, Marcus, Tang, Chan, Zhen, Reizenstein, Teboul, Zhong, Jin,
  Yang, Cummings, Carvill, Shepard, McPhie, Torres, Ginsburg, Wang, Wu, U,
  Saxena, Khandelwal, Zand, Matosich, Veeraraghavan, Michelena, Li, Jagadeesh,
  Huang, Chawla, Huang, Chen, Garg, A, Silva, Bell, Zhang, Guo, Yu, Moshkovich,
  Wehrstedt, Khabsa, Avalani, Bhatt, Mankus, Hasson, Lennie, Reso, Groshev,
  Naumov, Lathi, Keneally, Liu, Seltzer, Valko, Restrepo, Patel, Vyatskov,
  Samvelyan, Clark, Macey, Wang, Hermoso, Metanat, Rastegari, Bansal,
  Santhanam, Parks, White, Bawa, Singhal, Egebo, Usunier, Mehta, Laptev, Dong,
  Cheng, Chernoguz, Hart, Salpekar, Kalinli, Kent, Parekh, Saab, Balaji,
  Rittner, Bontrager, Roux, Dollar, Zvyagina, Ratanchandani, Yuvraj, Liang,
  Alao, Rodriguez, Ayub, Murthy, Nayani, Mitra, Parthasarathy, Li, Hogan,
  Battey, Wang, Howes, Rinott, Mehta, Siby, Bondu, Datta, Chugh, Hunt, Dhillon,
  Sidorov, Pan, Mahajan, Verma, Yamamoto, Ramaswamy, Lindsay, Lindsay, Feng,
  Lin, Zha, Patil, Shankar, Zhang, Zhang, Wang, Agarwal, Sajuyigbe, Chintala,
  Max, Chen, Kehoe, Satterfield, Govindaprasad, Gupta, Deng, Cho, Virk,
  Subramanian, Choudhury, Goldman, Remez, Glaser, Best, Koehler, Robinson, Li,
  Zhang, Matthews, Chou, Shaked, Vontimitta, Ajayi, Montanez, Mohan, Kumar,
  Mangla, Ionescu, Poenaru, Mihailescu, Ivanov, Li, Wang, Jiang, Bouaziz,
  Constable, Tang, Wu, Wang, Wu, Gao, Kleinman, Chen, Hu, Jia, Qi, Li, Zhang,
  Zhang, Adi, Nam, Yu, Wang, Zhao, Hao, Qian, Li, He, Rait, DeVito, Rosnbrick,
  Wen, Yang, Zhao, and Ma]{llama3_8b}
Aaron Grattafiori, Abhimanyu Dubey, Abhinav Jauhri, Abhinav Pandey, Abhishek
  Kadian, Ahmad Al-Dahle, Aiesha Letman, Akhil Mathur, Alan Schelten, Alex
  Vaughan, Amy Yang, Angela Fan, Anirudh Goyal, Anthony Hartshorn, Aobo Yang,
  Archi Mitra, Archie Sravankumar, Artem Korenev, Arthur Hinsvark, Arun Rao,
  Aston Zhang, Aurelien Rodriguez, Austen Gregerson, Ava Spataru, Baptiste
  Roziere, Bethany Biron, Binh Tang, Bobbie Chern, Charlotte Caucheteux, Chaya
  Nayak, Chloe Bi, Chris Marra, Chris McConnell, Christian Keller, Christophe
  Touret, Chunyang Wu, Corinne Wong, Cristian~Canton Ferrer, Cyrus Nikolaidis,
  Damien Allonsius, Daniel Song, Danielle Pintz, Danny Livshits, Danny Wyatt,
  David Esiobu, Dhruv Choudhary, Dhruv Mahajan, Diego Garcia-Olano, Diego
  Perino, Dieuwke Hupkes, Egor Lakomkin, Ehab AlBadawy, Elina Lobanova, Emily
  Dinan, Eric~Michael Smith, Filip Radenovic, Francisco Guzmán, Frank Zhang,
  Gabriel Synnaeve, Gabrielle Lee, Georgia~Lewis Anderson, Govind Thattai,
  Graeme Nail, Gregoire Mialon, Guan Pang, Guillem Cucurell, Hailey Nguyen,
  Hannah Korevaar, Hu~Xu, Hugo Touvron, Iliyan Zarov, Imanol~Arrieta Ibarra,
  Isabel Kloumann, Ishan Misra, Ivan Evtimov, Jack Zhang, Jade Copet, Jaewon
  Lee, Jan Geffert, Jana Vranes, Jason Park, Jay Mahadeokar, Jeet Shah, Jelmer
  van~der Linde, Jennifer Billock, Jenny Hong, Jenya Lee, Jeremy Fu, Jianfeng
  Chi, Jianyu Huang, Jiawen Liu, Jie Wang, Jiecao Yu, Joanna Bitton, Joe
  Spisak, Jongsoo Park, Joseph Rocca, Joshua Johnstun, Joshua Saxe, Junteng
  Jia, Kalyan~Vasuden Alwala, Karthik Prasad, Kartikeya Upasani, Kate Plawiak,
  Ke~Li, Kenneth Heafield, Kevin Stone, Khalid El-Arini, Krithika Iyer, Kshitiz
  Malik, Kuenley Chiu, Kunal Bhalla, Kushal Lakhotia, Lauren Rantala-Yeary,
  Laurens van~der Maaten, Lawrence Chen, Liang Tan, Liz Jenkins, Louis Martin,
  Lovish Madaan, Lubo Malo, Lukas Blecher, Lukas Landzaat, Luke de~Oliveira,
  Madeline Muzzi, Mahesh Pasupuleti, Mannat Singh, Manohar Paluri, Marcin
  Kardas, Maria Tsimpoukelli, Mathew Oldham, Mathieu Rita, Maya Pavlova,
  Melanie Kambadur, Mike Lewis, Min Si, Mitesh~Kumar Singh, Mona Hassan, Naman
  Goyal, Narjes Torabi, Nikolay Bashlykov, Nikolay Bogoychev, Niladri
  Chatterji, Ning Zhang, Olivier Duchenne, Onur Çelebi, Patrick Alrassy,
  Pengchuan Zhang, Pengwei Li, Petar Vasic, Peter Weng, Prajjwal Bhargava,
  Pratik Dubal, Praveen Krishnan, Punit~Singh Koura, Puxin Xu, Qing He,
  Qingxiao Dong, Ragavan Srinivasan, Raj Ganapathy, Ramon Calderer,
  Ricardo~Silveira Cabral, Robert Stojnic, Roberta Raileanu, Rohan Maheswari,
  Rohit Girdhar, Rohit Patel, Romain Sauvestre, Ronnie Polidoro, Roshan
  Sumbaly, Ross Taylor, Ruan Silva, Rui Hou, Rui Wang, Saghar Hosseini, Sahana
  Chennabasappa, Sanjay Singh, Sean Bell, Seohyun~Sonia Kim, Sergey Edunov,
  Shaoliang Nie, Sharan Narang, Sharath Raparthy, Sheng Shen, Shengye Wan,
  Shruti Bhosale, Shun Zhang, Simon Vandenhende, Soumya Batra, Spencer Whitman,
  Sten Sootla, Stephane Collot, Suchin Gururangan, Sydney Borodinsky, Tamar
  Herman, Tara Fowler, Tarek Sheasha, Thomas Georgiou, Thomas Scialom, Tobias
  Speckbacher, Todor Mihaylov, Tong Xiao, Ujjwal Karn, Vedanuj Goswami, Vibhor
  Gupta, Vignesh Ramanathan, Viktor Kerkez, Vincent Gonguet, Virginie Do, Vish
  Vogeti, Vítor Albiero, Vladan Petrovic, Weiwei Chu, Wenhan Xiong, Wenyin Fu,
  Whitney Meers, Xavier Martinet, Xiaodong Wang, Xiaofang Wang, Xiaoqing~Ellen
  Tan, Xide Xia, Xinfeng Xie, Xuchao Jia, Xuewei Wang, Yaelle Goldschlag,
  Yashesh Gaur, Yasmine Babaei, Yi~Wen, Yiwen Song, Yuchen Zhang, Yue Li,
  Yuning Mao, Zacharie~Delpierre Coudert, Zheng Yan, Zhengxing Chen, Zoe
  Papakipos, Aaditya Singh, Aayushi Srivastava, Abha Jain, Adam Kelsey, Adam
  Shajnfeld, Adithya Gangidi, Adolfo Victoria, Ahuva Goldstand, Ajay Menon,
  Ajay Sharma, Alex Boesenberg, Alexei Baevski, Allie Feinstein, Amanda Kallet,
  Amit Sangani, Amos Teo, Anam Yunus, Andrei Lupu, Andres Alvarado, Andrew
  Caples, Andrew Gu, Andrew Ho, Andrew Poulton, Andrew Ryan, Ankit Ramchandani,
  Annie Dong, Annie Franco, Anuj Goyal, Aparajita Saraf, Arkabandhu Chowdhury,
  Ashley Gabriel, Ashwin Bharambe, Assaf Eisenman, Azadeh Yazdan, Beau James,
  Ben Maurer, Benjamin Leonhardi, Bernie Huang, Beth Loyd, Beto~De Paola,
  Bhargavi Paranjape, Bing Liu, Bo~Wu, Boyu Ni, Braden Hancock, Bram Wasti,
  Brandon Spence, Brani Stojkovic, Brian Gamido, Britt Montalvo, Carl Parker,
  Carly Burton, Catalina Mejia, Ce~Liu, Changhan Wang, Changkyu Kim, Chao Zhou,
  Chester Hu, Ching-Hsiang Chu, Chris Cai, Chris Tindal, Christoph
  Feichtenhofer, Cynthia Gao, Damon Civin, Dana Beaty, Daniel Kreymer, Daniel
  Li, David Adkins, David Xu, Davide Testuggine, Delia David, Devi Parikh,
  Diana Liskovich, Didem Foss, Dingkang Wang, Duc Le, Dustin Holland, Edward
  Dowling, Eissa Jamil, Elaine Montgomery, Eleonora Presani, Emily Hahn, Emily
  Wood, Eric-Tuan Le, Erik Brinkman, Esteban Arcaute, Evan Dunbar, Evan
  Smothers, Fei Sun, Felix Kreuk, Feng Tian, Filippos Kokkinos, Firat Ozgenel,
  Francesco Caggioni, Frank Kanayet, Frank Seide, Gabriela~Medina Florez,
  Gabriella Schwarz, Gada Badeer, Georgia Swee, Gil Halpern, Grant Herman,
  Grigory Sizov, Guangyi, Zhang, Guna Lakshminarayanan, Hakan Inan, Hamid
  Shojanazeri, Han Zou, Hannah Wang, Hanwen Zha, Haroun Habeeb, Harrison
  Rudolph, Helen Suk, Henry Aspegren, Hunter Goldman, Hongyuan Zhan, Ibrahim
  Damlaj, Igor Molybog, Igor Tufanov, Ilias Leontiadis, Irina-Elena Veliche,
  Itai Gat, Jake Weissman, James Geboski, James Kohli, Janice Lam, Japhet
  Asher, Jean-Baptiste Gaya, Jeff Marcus, Jeff Tang, Jennifer Chan, Jenny Zhen,
  Jeremy Reizenstein, Jeremy Teboul, Jessica Zhong, Jian Jin, Jingyi Yang, Joe
  Cummings, Jon Carvill, Jon Shepard, Jonathan McPhie, Jonathan Torres, Josh
  Ginsburg, Junjie Wang, Kai Wu, Kam~Hou U, Karan Saxena, Kartikay Khandelwal,
  Katayoun Zand, Kathy Matosich, Kaushik Veeraraghavan, Kelly Michelena, Keqian
  Li, Kiran Jagadeesh, Kun Huang, Kunal Chawla, Kyle Huang, Lailin Chen,
  Lakshya Garg, Lavender A, Leandro Silva, Lee Bell, Lei Zhang, Liangpeng Guo,
  Licheng Yu, Liron Moshkovich, Luca Wehrstedt, Madian Khabsa, Manav Avalani,
  Manish Bhatt, Martynas Mankus, Matan Hasson, Matthew Lennie, Matthias Reso,
  Maxim Groshev, Maxim Naumov, Maya Lathi, Meghan Keneally, Miao Liu,
  Michael~L. Seltzer, Michal Valko, Michelle Restrepo, Mihir Patel, Mik
  Vyatskov, Mikayel Samvelyan, Mike Clark, Mike Macey, Mike Wang, Miquel~Jubert
  Hermoso, Mo~Metanat, Mohammad Rastegari, Munish Bansal, Nandhini Santhanam,
  Natascha Parks, Natasha White, Navyata Bawa, Nayan Singhal, Nick Egebo,
  Nicolas Usunier, Nikhil Mehta, Nikolay~Pavlovich Laptev, Ning Dong, Norman
  Cheng, Oleg Chernoguz, Olivia Hart, Omkar Salpekar, Ozlem Kalinli, Parkin
  Kent, Parth Parekh, Paul Saab, Pavan Balaji, Pedro Rittner, Philip Bontrager,
  Pierre Roux, Piotr Dollar, Polina Zvyagina, Prashant Ratanchandani, Pritish
  Yuvraj, Qian Liang, Rachad Alao, Rachel Rodriguez, Rafi Ayub, Raghotham
  Murthy, Raghu Nayani, Rahul Mitra, Rangaprabhu Parthasarathy, Raymond Li,
  Rebekkah Hogan, Robin Battey, Rocky Wang, Russ Howes, Ruty Rinott, Sachin
  Mehta, Sachin Siby, Sai~Jayesh Bondu, Samyak Datta, Sara Chugh, Sara Hunt,
  Sargun Dhillon, Sasha Sidorov, Satadru Pan, Saurabh Mahajan, Saurabh Verma,
  Seiji Yamamoto, Sharadh Ramaswamy, Shaun Lindsay, Shaun Lindsay, Sheng Feng,
  Shenghao Lin, Shengxin~Cindy Zha, Shishir Patil, Shiva Shankar, Shuqiang
  Zhang, Shuqiang Zhang, Sinong Wang, Sneha Agarwal, Soji Sajuyigbe, Soumith
  Chintala, Stephanie Max, Stephen Chen, Steve Kehoe, Steve Satterfield,
  Sudarshan Govindaprasad, Sumit Gupta, Summer Deng, Sungmin Cho, Sunny Virk,
  Suraj Subramanian, Sy~Choudhury, Sydney Goldman, Tal Remez, Tamar Glaser,
  Tamara Best, Thilo Koehler, Thomas Robinson, Tianhe Li, Tianjun Zhang, Tim
  Matthews, Timothy Chou, Tzook Shaked, Varun Vontimitta, Victoria Ajayi,
  Victoria Montanez, Vijai Mohan, Vinay~Satish Kumar, Vishal Mangla, Vlad
  Ionescu, Vlad Poenaru, Vlad~Tiberiu Mihailescu, Vladimir Ivanov, Wei Li,
  Wenchen Wang, Wenwen Jiang, Wes Bouaziz, Will Constable, Xiaocheng Tang,
  Xiaojian Wu, Xiaolan Wang, Xilun Wu, Xinbo Gao, Yaniv Kleinman, Yanjun Chen,
  Ye~Hu, Ye~Jia, Ye~Qi, Yenda Li, Yilin Zhang, Ying Zhang, Yossi Adi, Youngjin
  Nam, Yu, Wang, Yu~Zhao, Yuchen Hao, Yundi Qian, Yunlu Li, Yuzi He, Zach Rait,
  Zachary DeVito, Zef Rosnbrick, Zhaoduo Wen, Zhenyu Yang, Zhiwei Zhao, and
  Zhiyu Ma.
\newblock The llama 3 herd of models, 2024.
\newblock URL \url{https://arxiv.org/abs/2407.21783}.

\bibitem[Griffiths \& Hern{\'a}ndez-Lobato(2020)Griffiths and
  Hern{\'a}ndez-Lobato]{griffiths2020constrained}
Ryan-Rhys Griffiths and Jos{\'e}~Miguel Hern{\'a}ndez-Lobato.
\newblock Constrained bayesian optimization for automatic chemical design using
  variational autoencoders.
\newblock \emph{Chemical science}, 11\penalty0 (2):\penalty0 577--586, 2020.

\bibitem[Ilharco et~al.(2022)Ilharco, Ribeiro, Wortsman, Gururangan, Schmidt,
  Hajishirzi, and Farhadi]{ilharco2022editing}
Gabriel Ilharco, Marco~Tulio Ribeiro, Mitchell Wortsman, Suchin Gururangan,
  Ludwig Schmidt, Hannaneh Hajishirzi, and Ali Farhadi.
\newblock Editing models with task arithmetic.
\newblock \emph{arXiv preprint arXiv:2212.04089}, 2022.

\bibitem[Jang et~al.(2024)Jang, Lee, Kim, and Lee]{jang2024model}
Chaeyun Jang, Hyungi Lee, Jungtaek Kim, and Juho Lee.
\newblock Model fusion through bayesian optimization in language model
  fine-tuning.
\newblock \emph{Advances in Neural Information Processing Systems},
  37:\penalty0 29878--29912, 2024.

\bibitem[Kim et~al.(2025)Kim, Gouk, Kim, and Hospedales]{kim2025model}
Taehoon Kim, Henry Gouk, Minyoung Kim, and Timothy Hospedales.
\newblock Model merging is secretly certifiable: Non-vacuous generalisation
  bounds for low-shot learning.
\newblock \emph{arXiv preprint arXiv:2505.15798}, 2025.

\bibitem[Knowles(2006)]{knowles2006parego}
Joshua Knowles.
\newblock Parego: A hybrid algorithm with on-line landscape approximation for
  expensive multiobjective optimization problems.
\newblock \emph{IEEE transactions on evolutionary computation}, 10\penalty0
  (1):\penalty0 50--66, 2006.

\bibitem[Li et~al.(2025{\natexlab{a}})Li, Di, Yang, Qian, Yang, Hao, Tang, and
  Zhou]{li2025s}
Bingdong Li, Zixiang Di, Yanting Yang, Hong Qian, Peng Yang, Hao Hao, Ke~Tang,
  and Aimin Zhou.
\newblock It’s morphing time: Unleashing the potential of multiple llms via
  multi-objective optimization.
\newblock \emph{IEEE Transactions on Evolutionary Computation},
  2025{\natexlab{a}}.

\bibitem[Li et~al.(2024)Li, Zhang, Bu, Wang, He, Fu, Wu, Bian, Chen, and
  Bengio]{li2024map}
Lu~Li, Tianyu Zhang, Zhiqi Bu, Suyuchen Wang, Huan He, Jie Fu, Yonghui Wu,
  Jiang Bian, Yong Chen, and Yoshua Bengio.
\newblock Map: Low-compute model merging with amortized pareto fronts via
  quadratic approximation.
\newblock \emph{arXiv preprint arXiv:2406.07529}, 2024.

\bibitem[Li et~al.(2025{\natexlab{b}})Li, Lu, Dai, Huang, Ding, and
  Lu]{li2025became}
Mei Li, Yuxiang Lu, Qinyan Dai, Suizhi Huang, Yue Ding, and Hongtao Lu.
\newblock Became: Bayesian continual learning with adaptive model merging.
\newblock \emph{arXiv preprint arXiv:2504.02666}, 2025{\natexlab{b}}.

\bibitem[Liu et~al.(2024)Liu, Wang, Wang, Chen, Li, Tu, Chu, Li, and
  Sui]{liu2024checkpoint}
Deyuan Liu, Zecheng Wang, Bingning Wang, Weipeng Chen, Chunshan Li, Zhiying Tu,
  Dianhui Chu, Bo~Li, and Dianbo Sui.
\newblock Checkpoint merging via bayesian optimization in llm pretraining.
\newblock \emph{arXiv preprint arXiv:2403.19390}, 2024.

\bibitem[Mathern et~al.(2021)Mathern, Steinholtz, Sj{\"o}berg, {\"O}nnheim, Ek,
  Rempling, Gustavsson, and Jirstrand]{mathern2021multi}
Alexandre Mathern, Olof~Skogby Steinholtz, Anders Sj{\"o}berg, Magnus
  {\"O}nnheim, Kristine Ek, Rasmus Rempling, Emil Gustavsson, and Mats
  Jirstrand.
\newblock Multi-objective constrained bayesian optimization for structural
  design.
\newblock \emph{Structural and Multidisciplinary Optimization}, 63\penalty0
  (2):\penalty0 689--701, 2021.

\bibitem[Shoemake(1985)]{shoemake1985animating}
Ken Shoemake.
\newblock Animating rotation with quaternion curves.
\newblock In \emph{Proceedings of the 12th annual conference on Computer
  graphics and interactive techniques}, pp.\  245--254, 1985.

\bibitem[Snoek et~al.(2012)Snoek, Larochelle, and Adams]{snoek2012practical}
Jasper Snoek, Hugo Larochelle, and Ryan~P Adams.
\newblock Practical bayesian optimization of machine learning algorithms.
\newblock \emph{Advances in neural information processing systems}, 25, 2012.

\bibitem[Team(2025)]{qwen3_4b}
Qwen Team.
\newblock Qwen3 technical report, 2025.
\newblock URL \url{https://arxiv.org/abs/2505.09388}.

\bibitem[Wortsman et~al.(2022)Wortsman, Ilharco, Gadre, Roelofs, Gontijo-Lopes,
  Morcos, Namkoong, Farhadi, Carmon, Kornblith, et~al.]{wortsman2022model}
Mitchell Wortsman, Gabriel Ilharco, Samir~Ya Gadre, Rebecca Roelofs, Raphael
  Gontijo-Lopes, Ari~S Morcos, Hongseok Namkoong, Ali Farhadi, Yair Carmon,
  Simon Kornblith, et~al.
\newblock Model soups: averaging weights of multiple fine-tuned models improves
  accuracy without increasing inference time.
\newblock In \emph{International conference on machine learning}, pp.\
  23965--23998. PMLR, 2022.

\bibitem[Wu et~al.(2020)Wu, Toscano-Palmerin, Frazier, and
  Wilson]{wu2020practical}
Jian Wu, Saul Toscano-Palmerin, Peter~I Frazier, and Andrew~Gordon Wilson.
\newblock Practical multi-fidelity bayesian optimization for hyperparameter
  tuning.
\newblock In \emph{Uncertainty in Artificial Intelligence}, pp.\  788--798.
  PMLR, 2020.

\bibitem[Yadav et~al.(2023)Yadav, Tam, Choshen, Raffel, and
  Bansal]{yadav2023ties}
Prateek Yadav, Derek Tam, Leshem Choshen, Colin~A Raffel, and Mohit Bansal.
\newblock Ties-merging: Resolving interference when merging models.
\newblock \emph{Advances in Neural Information Processing Systems},
  36:\penalty0 7093--7115, 2023.

\bibitem[Yang et~al.(2024)Yang, Shen, Guo, Wang, Cao, Zhang, and
  Tao]{yang2024model}
Enneng Yang, Li~Shen, Guibing Guo, Xingwei Wang, Xiaochun Cao, Jie Zhang, and
  Dacheng Tao.
\newblock Model merging in llms, mllms, and beyond: Methods, theories,
  applications, and opportunities.
\newblock \emph{ACM Computing Surveys}, 2024.

\bibitem[Yu et~al.(2024)Yu, Yu, Yu, Huang, and Li]{yu2024language}
Le~Yu, Bowen Yu, Haiyang Yu, Fei Huang, and Yongbin Li.
\newblock Language models are super mario: Absorbing abilities from homologous
  models as a free lunch.
\newblock In \emph{Forty-first International Conference on Machine Learning},
  2024.

\bibitem[Yuce \& Amasyali(2025)Yuce and Amasyali]{yuce2025kafa}
Muzaffer~Kaan Yuce and Mehmet~Fatih Amasyali.
\newblock Kafa-merge: Model merging method with layer based bayes+ linear
  search.
\newblock \emph{IEEE Access}, 2025.

\bibitem[Zhou et~al.(2023)Zhou, Lu, Mishra, Brahma, Basu, Luan, Zhou, and
  Hou]{ifeval}
Jeffrey Zhou, Tianjian Lu, Swaroop Mishra, Siddhartha Brahma, Sujoy Basu,
  Yi~Luan, Denny Zhou, and Le~Hou.
\newblock Instruction-following evaluation for large language models.
\newblock \emph{arXiv preprint arXiv:2311.07911}, 2023.

\end{thebibliography}
\bibliographystyle{tmlr}

\appendix

\section{Appendix}

\subsection{Ablation on Acquisition Function Choice}
\label{app:qnparego}
\begin{figure}[t]
  \centering
  \setlength{\tabcolsep}{1pt}
  \renewcommand{\arraystretch}{1.0}
  \begin{subfigure}[t]{0.30\textwidth}
    \centering
    \includegraphics[width=\linewidth,page=1]{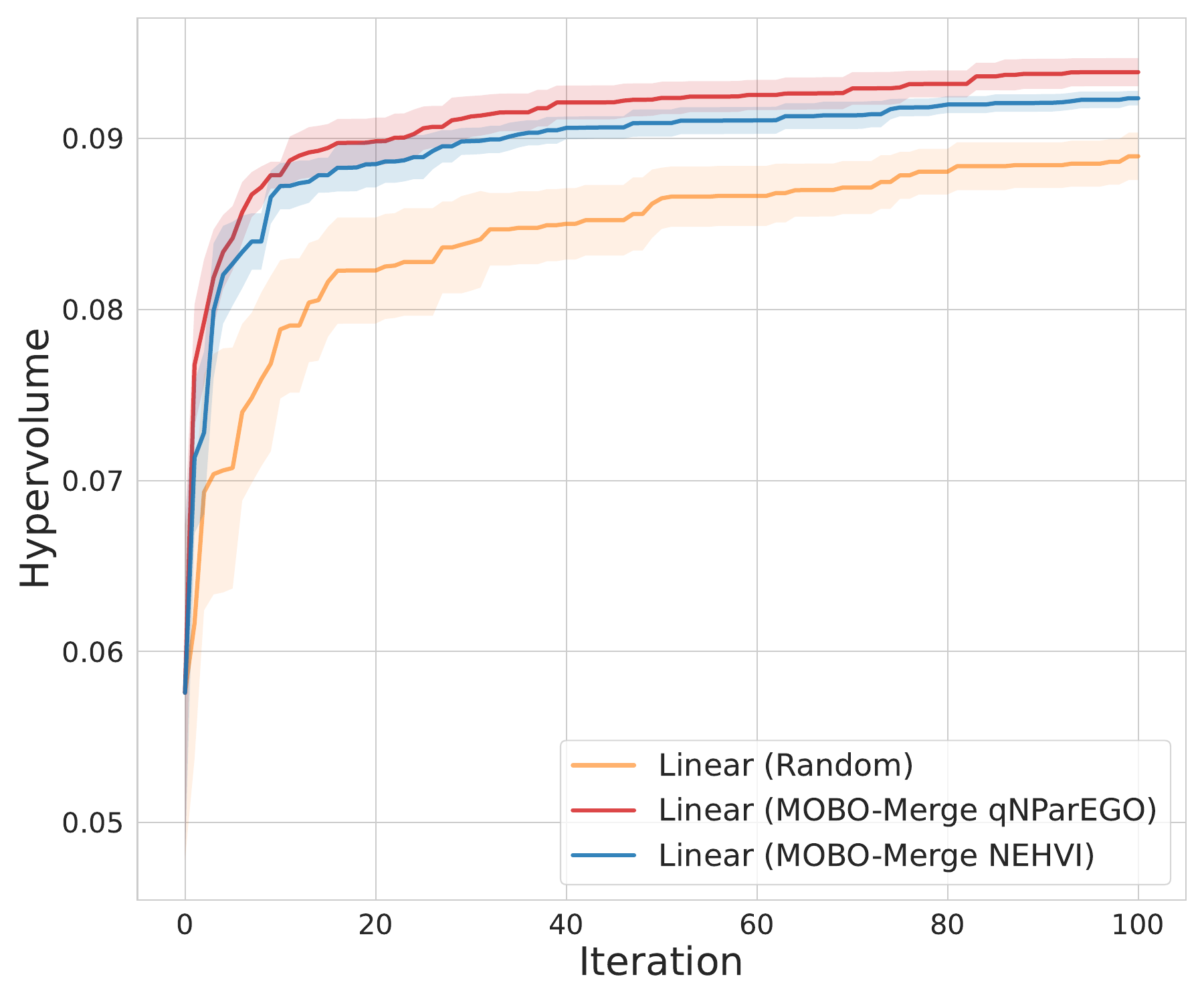}
    \caption{Linear}
  \end{subfigure}\hfill
  \begin{subfigure}[t]{0.30\textwidth}
    \centering
    \includegraphics[width=\linewidth,page=1]{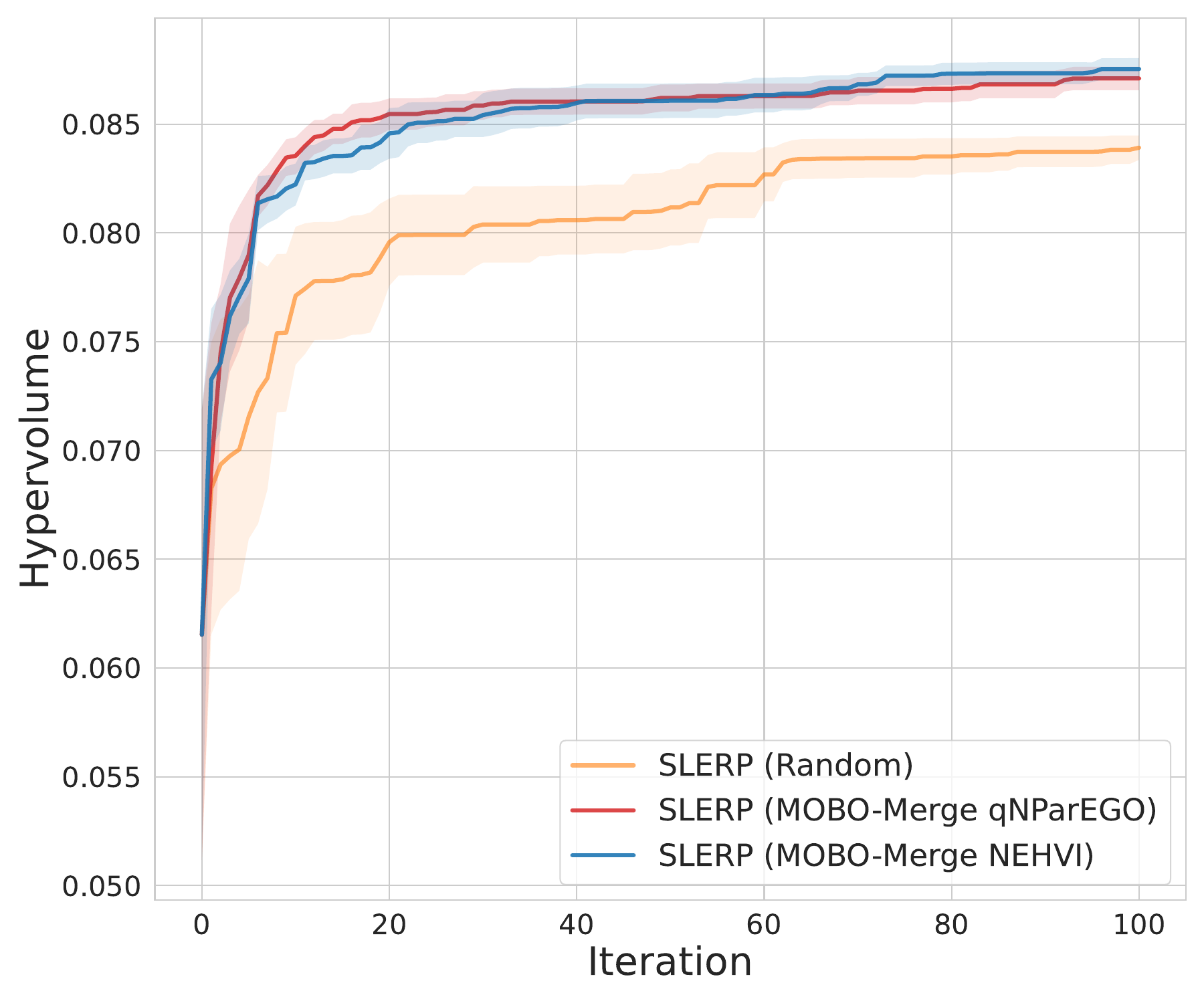}
    \caption{SLERP}
  \end{subfigure}\hfill
  \begin{subfigure}[t]{0.30\textwidth}
    \centering
    \includegraphics[width=\linewidth,page=1]{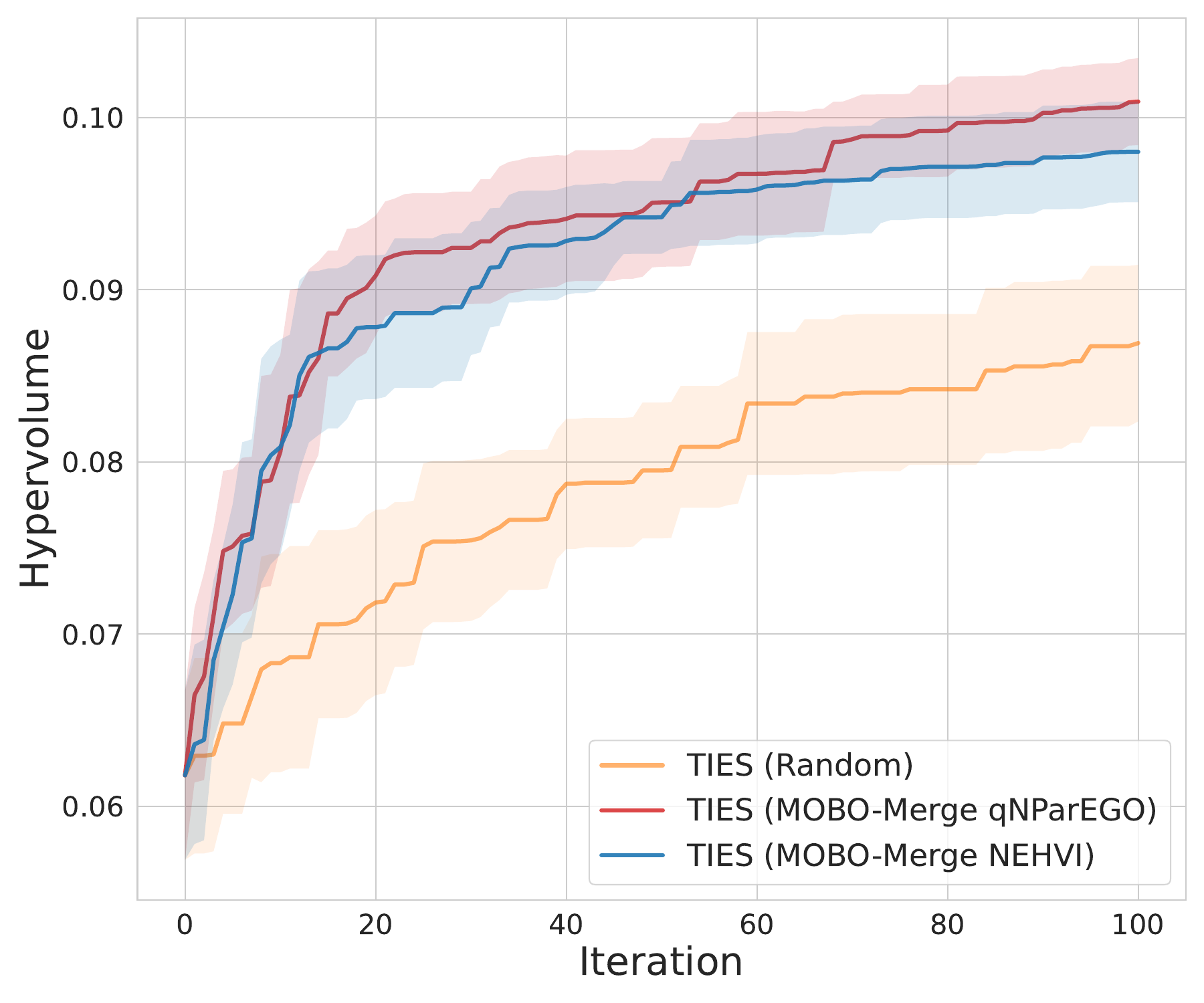}
    \caption{TIES}
  \end{subfigure}

  \vspace{0.2cm}

  \begin{subfigure}[t]{0.30\textwidth}
    \centering
    \includegraphics[width=\linewidth,page=1]{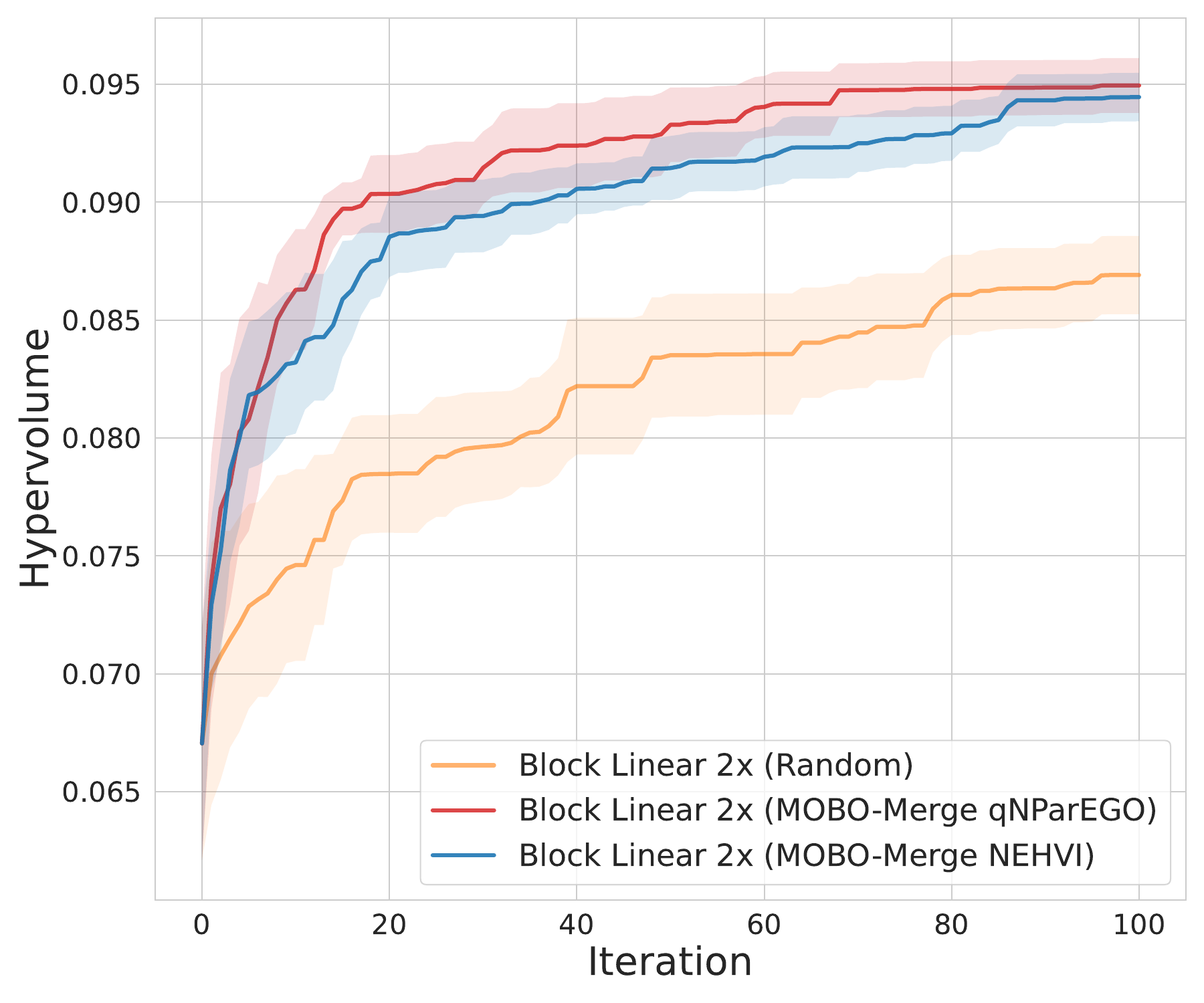}
    \caption{Block Linear 2x}
  \end{subfigure}\hfill
  \begin{subfigure}[t]{0.30\textwidth}
    \centering
    \includegraphics[width=\linewidth,page=1]{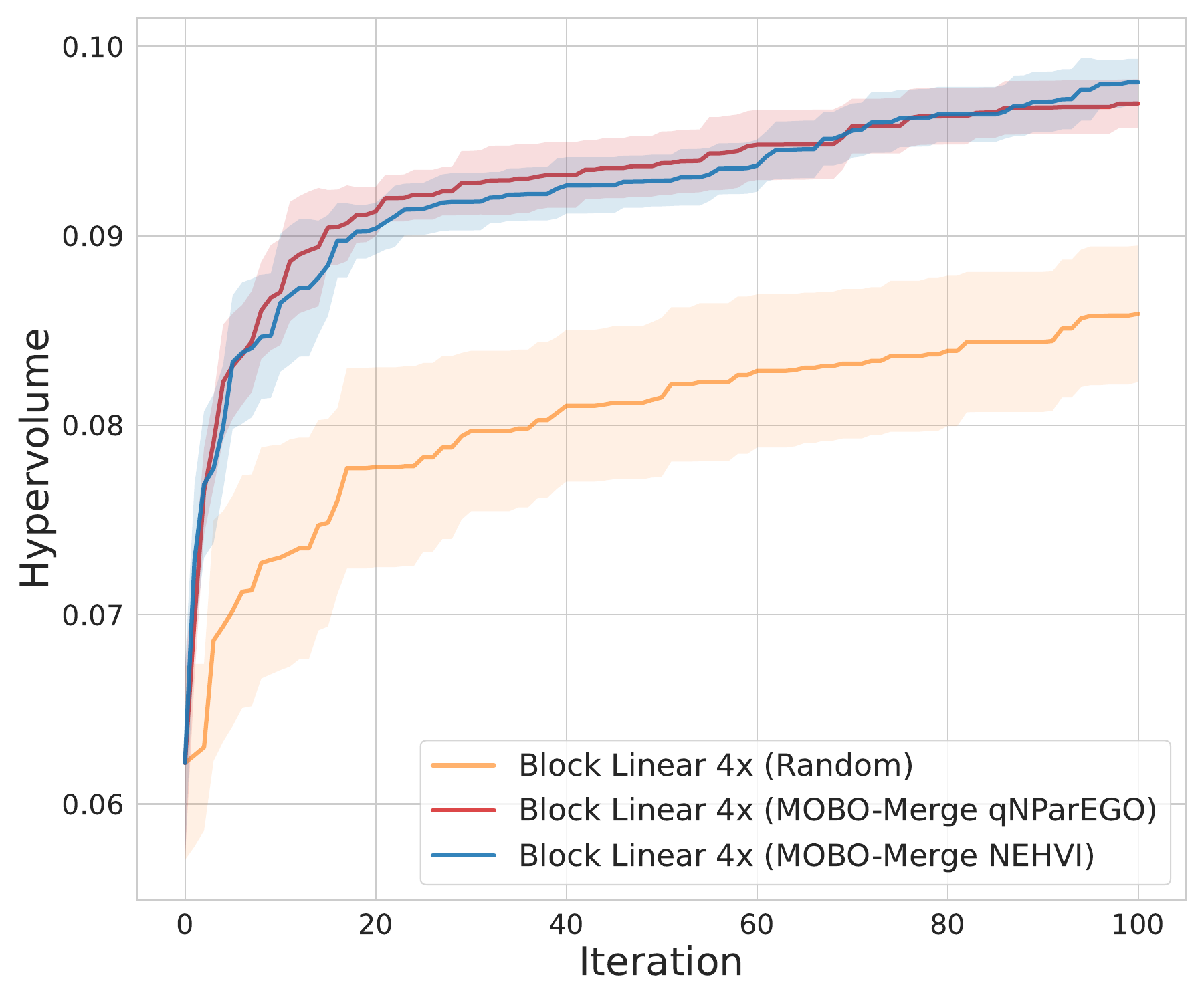}
    \caption{Block Linear 4x}
  \end{subfigure}\hfill
  \begin{subfigure}[t]{0.30\textwidth}
    \centering
    \includegraphics[width=\linewidth,page=1]{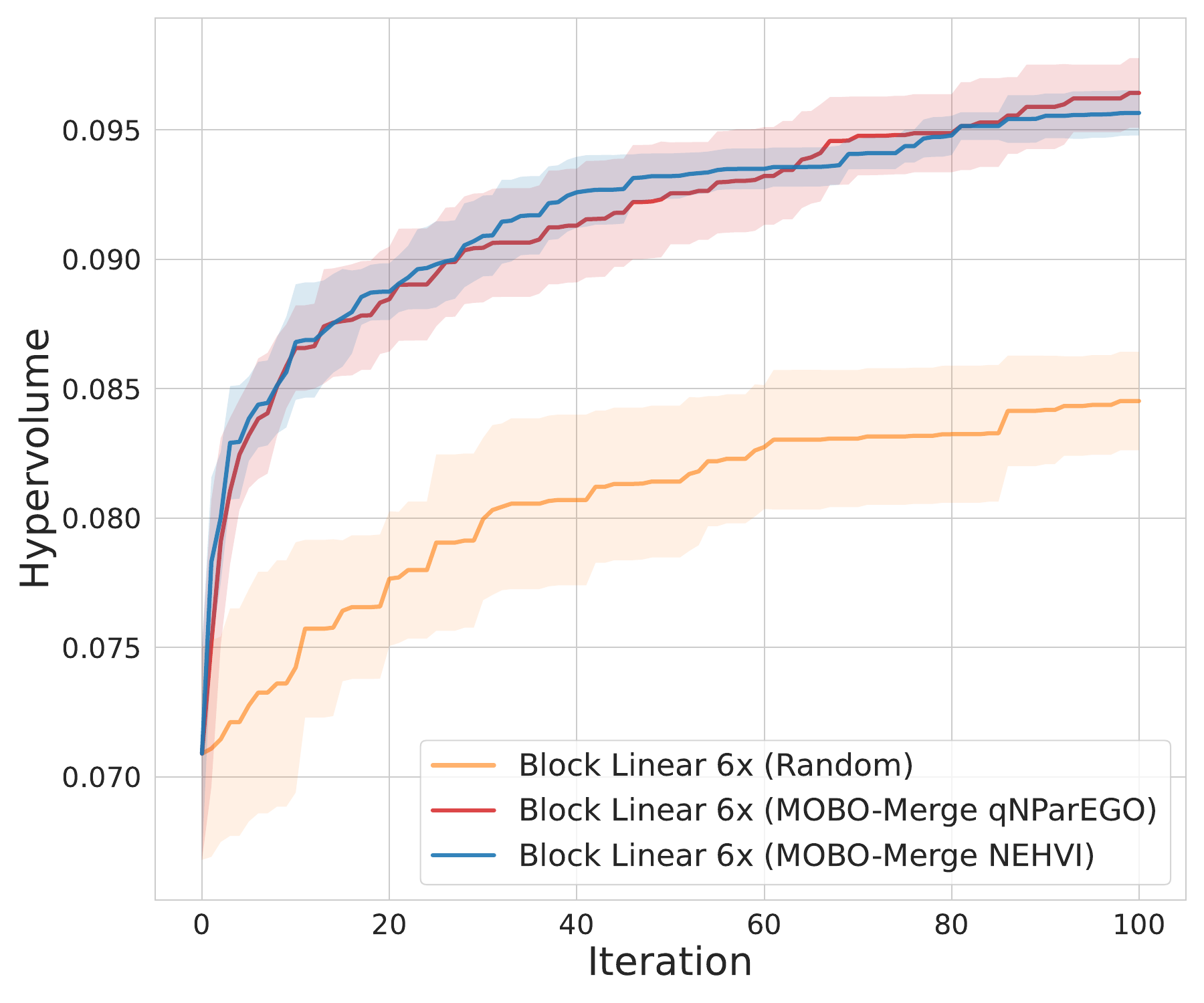}
    \caption{Block Linear 6x}
  \end{subfigure}

  \caption{\textbf{Acquisition-function ablation on Qwen3-4B AB.} Validation hypervolume over cumulative evaluations for NEHVI, qNParEGO, and random search across six merge operators. Curves show the mean across ten seeds, and shaded regions indicate the mean
$\pm$ one standard error. Neither acquisition function uniformly dominates.}
  \label{fig:nehvi-vs-nparego}
\end{figure}

We compare NEHVI with qNParEGO\citep{daulton2021parallel} on the Qwen AB setting. There is no uniform winner. qNParEGO has the higher final mean validation hypervolume for Linear, TIES, Block-Linear 2x, and Block-Linear 6x, while NEHVI is higher for SLERP and Block-Linear 4x. The absolute NEHVI-qNParEGO gaps range from 0.0004 to 0.0029, and both guided methods generally outperform random search. Thus, the ablation supports the robustness of the MOBO formulation to the acquisition choice. The results are shown in Figure~\ref{fig:nehvi-vs-nparego}.



\subsection{Models}
\label{app:modellinks}
We use two sets of architecture-compatible checkpoints. In each family, the base checkpoint is used as the shared reference model for TIES and is not an additional source in the AB or ABC merge. Models A, B, and C denote the source checkpoints associated with the three experimental objectives.

\paragraph{Qwen3-4B.}
\begin{itemize}
    \item \textbf{Base/TIES reference:}
    \texttt{Qwen/Qwen3-4B-Base}.
    \item \textbf{Model A (instruction following):}
    \texttt{Qwen/Qwen3-4B-Instruct-2507}.
    \item \textbf{Model B (mathematical reasoning):}
    \texttt{prithivMLmods/Draconis-Qwen3\_Math-4B-Preview}.
    \item \textbf{Model C (code generation):}
    \texttt{OpenHands/CodeScout-4B}.
\end{itemize}

\paragraph{Llama-3.1-8B.}
\begin{itemize}
    \item \textbf{Base/TIES reference:}
    \texttt{meta-llama/Llama-3.1-8B}.
    \item \textbf{Model A (instruction following):}
    \texttt{meta-llama/Llama-3.1-8B-Instruct}.
    \item \textbf{Model B (mathematical reasoning):}
    \texttt{nvidia/OpenMath2-Llama3.1-8B}.
    \item \textbf{Model C (code generation):}
    \texttt{tokyotech-llm/Llama-3.1-Swallow-8B-v0.5}.
\end{itemize}

The AB experiments merge Models A and B, while the ABC experiments add Model C. These labels describe each checkpoint's role in our experiments and the corresponding evaluation objective; in particular, the Llama Swallow checkpoint is a broader continually pretrained model rather than a code-only fine-tune.



\subsection{Compute Cost}
\label{app:compute}
Each candidate requires materializing an FP16 merged checkpoint and evaluating it on two tasks in AB or three tasks in ABC. All experiments were run on single RTX 6000 Ada GPUs with vLLM-backed evaluation. Representative end-to-end Linear runs required approximately 4 minutes per Qwen AB candidate, 6.5 minutes per Qwen ABC candidate, 6 minutes per Llama AB candidate, and 9 minutes per Llama ABC candidate. Runtime varies with the merge operator, generation lengths, and hardware, and held-out re-evaluation of the selected Pareto sets incurs additional cost.

\subsection{Reproducibility}
We use search seeds 42-51 and the same deterministic benchmark partition (split seed 4203) for every model, operator, and search method. Decoding uses temperature zero. Hypervolume curves report across-seed aggregates; the held-out table reports the mean across all 10 completed complementary-split re-evaluations. Configurations, per-candidate objective values, and
optimization-to-held-out mappings have been retained and will be released,
together with the code and remaining experimental artifacts, upon
publication.

\end{document}